\documentclass[10pt,journal,compsoc]{IEEEtran}
\ifCLASSOPTIONcompsoc
  \usepackage[nocompress]{cite}
\else
  \usepackage{cite}
\fi
\ifCLASSINFOpdf
\else
\fi
\usepackage{hyperref}       
\hypersetup{colorlinks=true}
\usepackage{url}            
\usepackage{booktabs}       
\usepackage{amsfonts}       
\usepackage{nicefrac}       
\usepackage{microtype}      
\usepackage{xcolor}         
\usepackage{amsmath}
\usepackage{multirow}
\usepackage{float}
\usepackage{color, colortbl}
\usepackage{xcolor}
\usepackage{rotating}
\usepackage{graphicx,wrapfig}
\usepackage[export]{adjustbox}
\usepackage{enumitem}
\usepackage[noend]{algorithm2e}
\usepackage{pifont}

\definecolor{dgreen}{RGB}{1,150,74}

\newcommand\red[1]{\textcolor{red}{#1}}
\newcommand\dgreen[1]{\textcolor{dgreen}{#1}}
\newcommand\gray[1]{\textcolor{gray}{#1}}
\newcommand\cyan[1]{\textcolor{cyan}{#1}}

\newcommand\up[1]{\textcolor{dgreen}{$\uparrow{#1}$}}

\newcommand\down[1]{\textcolor{red}{$\downarrow{#1}$}}
\newcommand\lighter[1]{\textcolor{dgreen}{$\downarrow{#1}$}}
\newcommand\mypar[1]{\par\vspace{-0mm}\noindent\textbf{#1}\;\;}
\newcommand\ut[1]{\textcolor{gray}{#1}} 

\newcommand{\cmark}{\ding{51}}%
\newcommand{\xmark}{\ding{55}}%
\newcommand\best[1]{\colorbox{cyan!20}{#1}}
\newcommand\bests[1]{\colorbox{gray!30}{#1}}

\def\MG{{ G}}
\def\MF{{ F}}
\def\MI{{ I}}
\def\MX{{ X}}

\def\ML{{ L}}

\begin{document}
%
\title{Super Sparse 3D Object Detection}
%
%
%
%

\author{Lue Fan,
        Yuxue Yang,
        Feng Wang,
        Naiyan Wang,
        and Zhaoxiang Zhang
\IEEEcompsocitemizethanks{
\IEEEcompsocthanksitem{Lue~Fan and Yuxue~Yang and Zhaoxiang~Zhang are with Center for Research on Intelligent Perception and Computing (CRIPAC), National Laboratory of Pattern Recognition (NLPR), Institute of Automation, Chinese Academy of Sciences (CASIA), Beijing 100190, China. E-mail: \{fanlue2019, yangyuxue2023, zhaoxiang.zhang\}@ia.ac.cn.}
\IEEEcompsocthanksitem{
          Feng~Wang and Naiyan~Wang are with TuSimple, Beijing 100020,
          China. E-mail: \{feng.wff, winsty\}@gmail.com.}
}
}

%
%

\markboth{Journal of \LaTeX\ Class Files,~Vol.~14, No.~8, August~2015}%
{Shell \MakeLowercase{\textit{et al.}}: Bare Demo of IEEEtran.cls for Computer Society Journals}
%



\IEEEtitleabstractindextext{%
\begin{abstract}
As the perception range of LiDAR expands, LiDAR-based 3D object detection contributes ever-increasingly to the long-range perception in autonomous driving.
Mainstream 3D object detectors often build dense feature maps, where the cost is quadratic to the perception range, making them hardly scale up to the long-range settings.
To enable efficient long-range detection, we first propose a fully sparse object detector termed FSD.
FSD is built upon the general sparse voxel encoder and a novel sparse instance recognition (SIR) module. 
SIR groups the points into instances and applies highly-efficient instance-wise feature extraction.
The instance-wise grouping sidesteps the issue of the center feature missing, which hinders the design of the fully sparse architecture.
To further enjoy the benefit of fully sparse characteristic, we leverage temporal information to remove data redundancy and propose a super sparse detector named FSD++.
FSD++ first generates residual points, which indicate the point changes between consecutive frames.
The residual points, along with a few previous foreground points, form the super sparse input data, greatly reducing data redundancy and computational overhead.
We comprehensively analyze our method on the large-scale Waymo Open Dataset, and state-of-the-art performance is reported.
To showcase the superiority of our method in long-range detection, we also conduct experiments on Argoverse 2 Dataset, where the perception range ($200m$) is much larger than Waymo Open Dataset ($75m$). 
Code is open-sourced at \url{https://github.com/tusen-ai/SST}.

\end{abstract}

}

\maketitle

\IEEEdisplaynontitleabstractindextext

%
\IEEEpeerreviewmaketitle

\begin{figure}

	\centering
	\includegraphics[width=0.98\columnwidth]{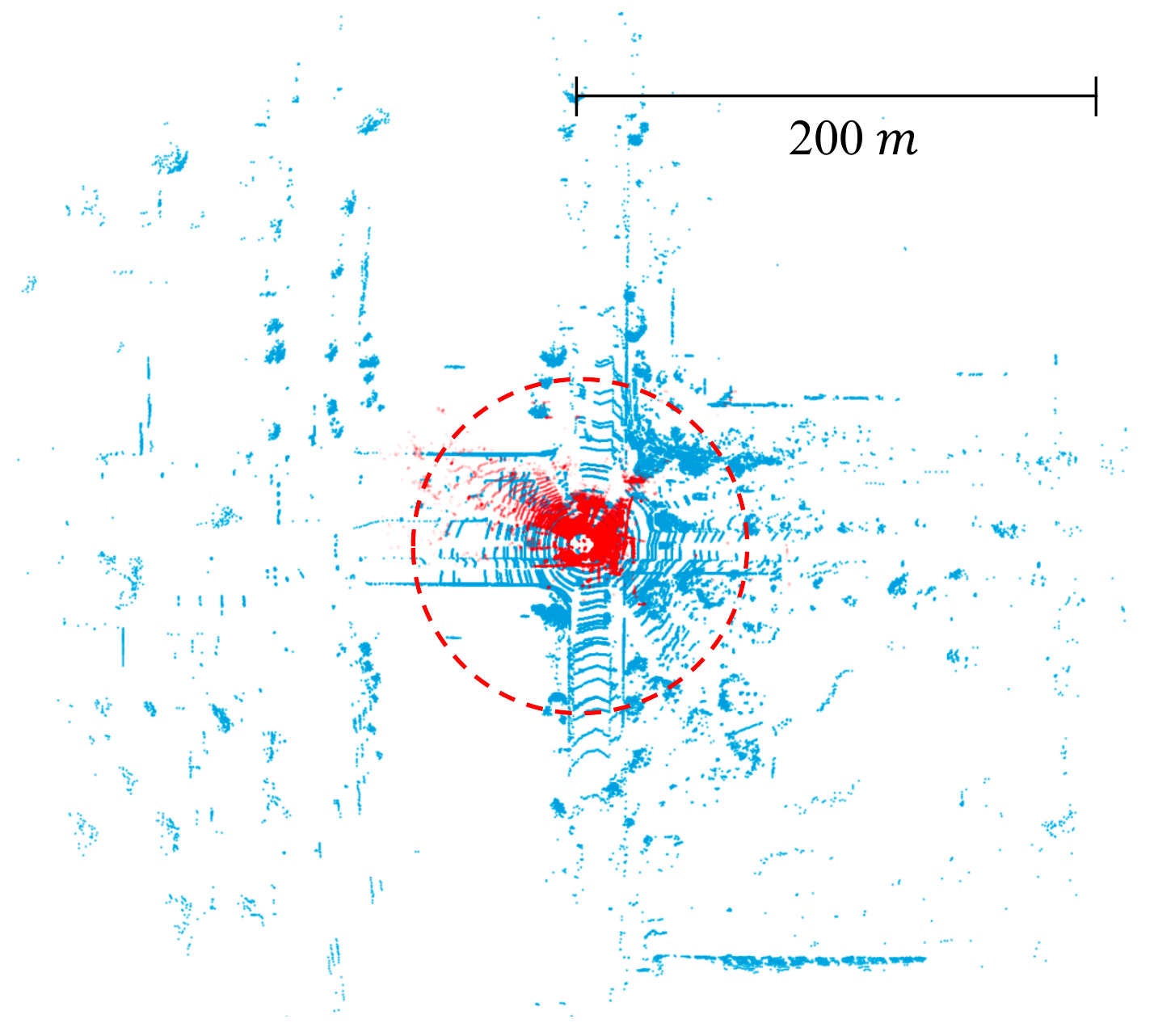}

	\caption{
	{Short-range point clouds (\red{red}, from KITTI~\cite{kitti}) v.s. long-range point clouds (\textcolor{cyan}{blue}, from Argoverse 2~\cite{argo2})}.
	The radius of the red circle is 75 meters.
	The sparsity quickly increases as the range extends.}

	\label{fig:argo_kitti}
\end{figure}

%
%
%
%
\IEEEraisesectionheading{\section{Introduction}\label{sec:intro}}
\IEEEPARstart{A}{utonomous} driving systems are eager for efficient long-range perception, especially in high-speed scenarios.
Current LiDAR-based 3D object detectors usually convert sparse features into dense feature maps for further feature extraction and prediction, which we name as \textbf{dense detectors}.
Dense detectors perform well on current popular benchmarks~\cite{WOD, kitti, nus}, where the perception range is relatively short (less than 75 meters).
However, it is impractical to scale the dense detectors to the long-range setting (more than 200 meters, Fig.~\ref{fig:argo_kitti}).
In such settings, the computational and spatial complexity on dense feature maps is quadratic to the perception range.
Fortunately, the sparsity of LiDAR point clouds also increases as the perception range extends (see Fig.~\ref{fig:argo_kitti}), and the calculation on the unoccupied area is essentially unnecessary.
Given the inherent sparsity, an essential solution for efficient long-range detection is to remove the dense feature maps and make the network architectures \textit{fully sparse}.
\par
\begin{figure}

	\centering
	\includegraphics[width=0.98\columnwidth]{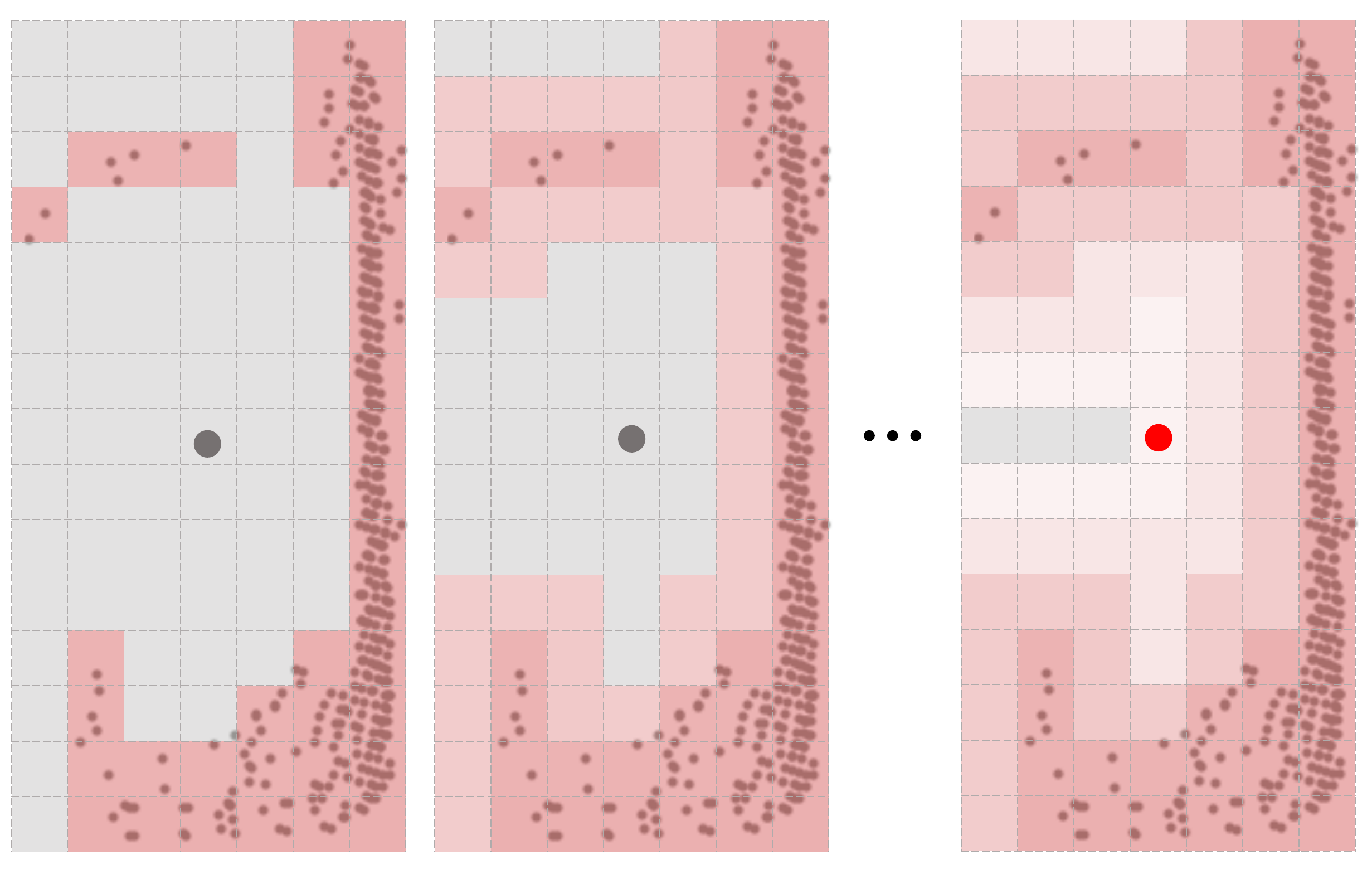}

	\caption{
	Illustration of \textbf{center feature missing} and \textbf{feature diffusion} on dense feature maps from Bird's Eye View.
	The empty instance center (red dot) is filled by the features diffused from occupied voxels (with LiDAR points), after several convolutions.
}
	\label{fig:diffusion}
\end{figure}

However, removing the dense feature map is non-trivial since it plays an indispensable role in current designs.
Commonly adopted sparse voxel encoders~\cite{second,sst,parta2} only extract the features on the non-empty voxels.
Without dense feature maps, the object centers are usually empty, especially for large objects. 
We name this issue as ``\textbf{Center Feature Missing (CFM)}'' (Fig.~\ref{fig:diffusion}).
CFM significantly weakens the representation power of the center voxels, even making the center feature empty in some extreme cases like super large vehicles.
However, almost all popular voxel or pillar based detectors~\cite{pvrcnn, sst, centerpoint, pvrcnnpp, second} adopt center-based assignment and rely on the center feature since it is an ideal representation of the whole object.
So they have to first convert sparse voxels to dense feature maps in Bird's Eye View after the sparse voxel encoder.
Then they resolve the CFM issue by applying convolutions on the dense feature maps to diffuse features to instance centers, which we name as \textbf{feature diffusion} (Fig.~\ref{fig:diffusion}).

To properly eliminate the dense feature map, we investigate the purely point-based detectors because they are naturally fully sparse.
However, two drawbacks limit the usage of point-based methods. 
(1) The time-consuming neighborhood query~\cite{pointnet++} is the long-standing difficulty to apply it to large-scale point cloud (more than 100K points).
(2) To reduce the computational overhead, point-based methods aggressively downsample the whole scene to a fixed number of points.
The aggressive downsampling leads to inevitable information loss and insufficient recall of foreground objects~\cite{3dssd, iassd}, especially for small ones.
As a result, very few purely point-based detectors have reached state-of-the-art performance in the recent benchmarks with large-scale point clouds.

In this paper, we first propose \textbf{Fully Sparse Detector (FSD)} to sidestep the issue of center feature missing.
FSD is built upon a general sparse voxel encoder ~\cite{second, parta2, sst} for voxel/point feature extraction.
Then FSD groups the points into an instance, and further extract the instance-level feature and predict a single bounding box from the integrated instance feature, via a novel \textbf{Sparse Instance Recognition (SIR)} module.
In this way, predictions are made from the whole instance feature instead of the weak or missed center feature.
As a point-based module, SIR has several desired properties:
(1) Unlike previous point-based modules, SIR simply treats instances as groups, and does not apply the time-consuming neighborhood query for further grouping.
(2) Similar to dynamic voxelization~\cite{MVF}, SIR leverages \emph{dynamic broadcast/pooling} for tensor manipulation to avoid point sampling or padding.
(3) Since the group in SIR covers the whole instance, it builds a sufficient receptive field regardless of the physical size of the instance.
\par
To unleash the full potential of FSD, we further utilize temporal information and propose a \textbf{Super Sparse 3D Object Detector}, named FSD++. FSD++ is inspired by human visual behavior: human is sensitive to and focuses on dynamic parts of the physical world.
In particular, FSD++ utilizes ego-motion to remove the static parts containing heavy temporal redundancy, while only retaining the informative dynamic parts.
We name the detected dynamic parts as residual points since the process is similar to applying the difference between frames.
In this way, we create a \emph{super sparse} point cloud consisting of residual points and a small number of past foreground points from history predictions. 
FSD++ then takes the super sparse point cloud as input, achieving a very efficient detection framework with temporal fusion.
We owe the credit of the high efficiency to the synergy of the fully sparse characteristic and the super sparse input.
We list our contributions as follows.
\begin{itemize}
    \item We introduce the concept of Fully Sparse Detector (FSD), which is the essential solution for efficient long-range LiDAR detection. We further propose Sparse Instance Recognition (SIR) to sidestep the issue of Center Feature Missing (CFM) in sparse feature maps. Combining SIR with general sparse voxel encoders, we develop an efficient and effective FSD implementation.
    \item Based on FSD, we further present the FSD++ framework, which aggregates a super sparse point cloud from multi-frames as input, yet removing the temporal redundancy of point clouds.
    The proposed framework uncovers the untapped potential of sparse architecture. We hope our efforts attract the attention of the community to fully sparse architecture.
    \item
    FSD achieves state-of-the-art performance on the competitive Waymo Open Dataset.
    Besides, we further apply our method to the recently released Argoverse 2 dataset to demonstrate the superiority of FSD in long-range detection, where FSD is much more efficient than its dense counterparts.
    FSD++ achieves comparable performance with mainstream state-of-the-art multi-frame detectors with minimal additional overhead compared with single-frame input.
\end{itemize}

\section{Related Work}
In reviewing the evolution of LiDAR-based 3D object detectors, the previous methods could be categorized into three types by their spatial sparsity: dense detectors, sparse detectors, and semi-dense detectors.
Below, we provide a brief revisit of previous arts according to spatial sparsity.
\subsection{Voxel-based Dense Detectors}
Pioneering work 3DFCN~\cite{3dfcn} and VoxelNet~\cite{voxelnet} use dense convolution for voxel feature extraction.
They bring convolutional neural networks to the field of LiDAR-based 3D object detection and achieve competitive results at the time.
However, it is inefficient to apply dense convolution to 3D voxel representation.
MV3D~\cite{mv3d}, PIXOR~\cite{pixor}, and PointPillars~\cite{pointpillar} adopt 2D dense convolution in Bird's Eye View (BEV) feature maps achieving significant efficiency improvement.
We refer to such detectors as \textbf{dense detectors} since they convert the sparse point cloud into dense feature maps.
\subsection{Point-based Sparse Detectors}
Since PointNet~\cite{pointnet} and PointNet++~\cite{pointnet++} shed light on the deep learning for 3D point sets, a series of point-based detectors have emerged.
These purely point-based detectors are born to be fully sparse.
PointRCNN~\cite{pointrcnn} is the pioneering work of this line of work.
3DSSD~\cite{3dssd} accelerates the point-based method by removing the feature propagation layer and refinement module.
VoteNet~\cite{votenet} first makes a center voting and then generates proposals from the voted center achieving better accuracy.
Albeit many methods~\cite{3dssd, iassd, dynamicball} have tried to accelerate the point-based method, the time-consuming point sampling and neighborhood query are still unaffordable in large-scale point clouds (more than 100k points per scene).
So current benchmarks~\cite{WOD, nus} with large-scale point clouds are still dominated by voxel-based dense/semi-dense detectors~\cite{afdetv2,pvrcnnpp,deepfusion}.

\subsection{Semi-dense Detectors}
Different from dense detectors, \textbf{semi-dense detectors} incorporate both sparse features and dense features.
SECOND~\cite{second} employs sparse convolution to extract the sparse voxel features in 3D space, which then are converted to dense feature maps in BEV to enlarge the receptive field and integrate with 2D detection head~\cite{ssd, fasterrcnn, centernet}.
Based on SECOND-style semi-dense detectors, a series of work~\cite{se-ssd, cia-ssd, sa-ssd} made further improvements on the single-stage paradigm.
And other methods attach a second stage for fine-grained feature extraction and proposal refinement~\cite{parta2, pvrcnn, pvrcnnpp, voxelrcnn}, achieving superior performance.
Although semi-dense detectors become dominating in academia and industry, related research has stagnated here because the semi-dense detector cannot be trivially lifted to be fully sparse as we discussed in Sec.~\ref{sec:intro}.

\section{FSD: Fully Sparse 3D Object Detection}
\subsection{Overall Architecture}
\begin{figure*}[t]
	\centering
	\includegraphics[width=0.99\linewidth]{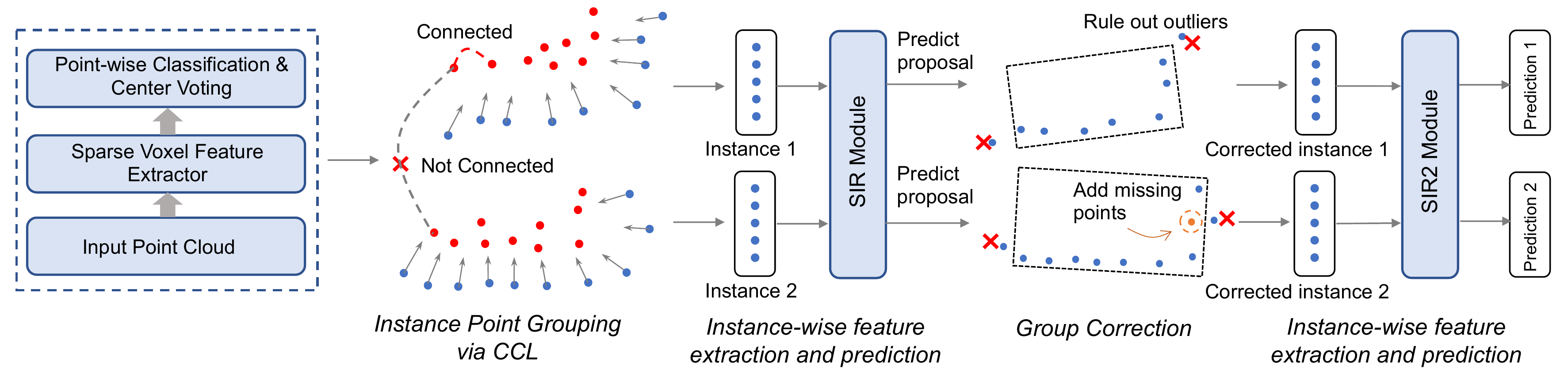}
	\caption{
	{Overall architecture of FSD.}
	For simplicity, we only use two instances to illustrate the pipeline. 
	\red{Red} dots are the voted centers from each LiDAR point (\textcolor{blue}{blue} dots).
	The SIR module and the SIR2 module all contain 3 SIR layers.  }
	\label{fig:overall}
\end{figure*}
Following the motivation of instances as groups, we have four steps to build the fully sparse detector (FSD):
1) We first utilize a sparse voxel encoder~\cite{sst, parta2, second} to extract voxel features and casts votes for object centers(Sec.~\ref{sec:grouping}).
2) {Instance Point Grouping} groups foreground points into instances based on the voting results (Sec.~\ref{sec:grouping}).
3) Given the grouping results, {Sparse Instance Recognition (SIR)} module extracts instance/point features and generates proposals (Sec.~\ref{sec:sir}). 
4) The proposals are utilized to correct the point grouping and refine the proposals iteratively (Sec.~\ref{sec:correction}).
\subsection{Instance Point Grouping}
\label{sec:grouping}
\mypar{Classification and Voting} We first extract voxel features from the point cloud with a sparse voxel encoder, such as sparse attention blocks in SST~\cite{sst} or sparse convolution encoder.
Then we build point features by concatenating voxel features and the offsets from points to their corresponding voxel centers.
These point features are passed into two heads for foreground classification and center voting.
The voting is similar to VoteNet~\cite{votenet}, where the model predicts the offsets from foreground points to corresponding object centers.
L1 loss~\cite{fasterrcnn} and Focal Loss~\cite{focalloss} are adopted as voting loss $\ML_{vote}$ and semantic classification loss $\ML_{sem}$.
\vspace{1mm}
\mypar{Connected Components Labeling (CCL)} 
To group points into instances, we regard all the predicted centers (red dots in Fig.~\ref{fig:overall}) as vertices in a graph.
Two vertices are connected if their distance is smaller than a certain threshold.
Then a connected component in this graph can be viewed as an instance, and all points voted to this connected component share a group ID.
Unlike the ball query in VoteNet, our CCL-based grouping avoids fragmented instances in most cases.
Although there are many elaborately designed instance grouping methods~\cite{pointgroup, wang2018sgpn, panopticcylinder}, we opt for the simple CCL because it is adequate in our design and can be implemented by the efficient Union-Find algorithm~\cite{tarjan1979class} in parallel.
\begin{figure*}
	\centering
	\includegraphics[width=0.99\linewidth]{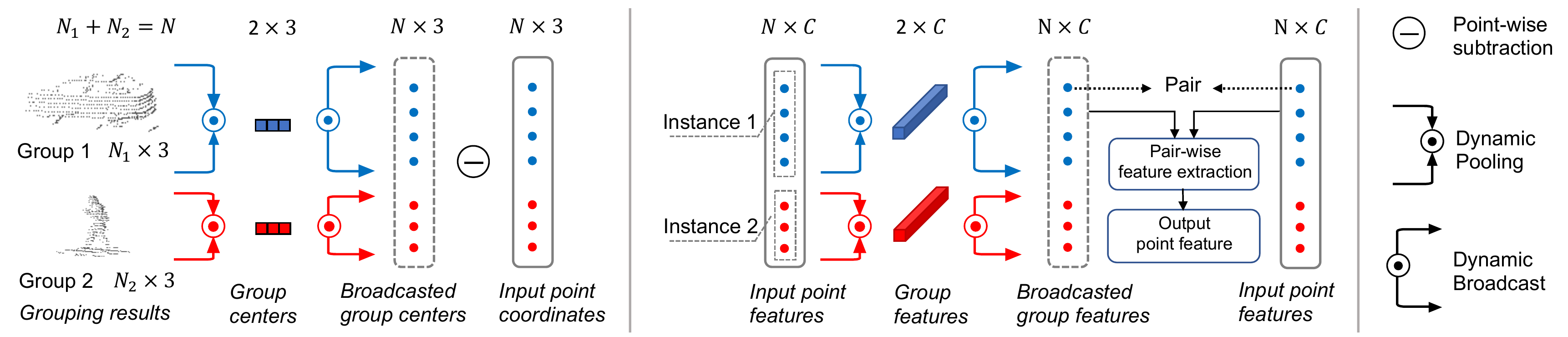}
	\caption{
	{Illustration of building instance-level point operators with dynamic broadcast/pooling.}
	Best viewed in color.
	Left: calculating center-to-neighbor offsets given raw point clouds.
	Right: updating point features. Note that the operation is parallel among all instances.
	}
	\label{fig:dynamic_op}
\end{figure*}
\subsection{Sparse Instance Recognition}
\label{sec:sir}
\subsubsection{Preliminaries: Dynamic Broadcast/Pooling}
\label{subsec:preliminary}
Given $\texttt{N}$ points belong to $\texttt{M}$ groups, we define their corresponding group ID array as $\MI$ in the shape of $[\texttt{N},]$ and their feature array as $\MF$ in the shape of $[\texttt{N,C}]$,
where $\texttt{C}$ is the feature dimensions.
$\MF^{(i)}$ is the feature array of points belonging to the $i$-th group.
Dynamic pooling aggregates each $\MF^{(i)}$ into one group feature $\boldsymbol{g}_i$ of shape $[\texttt{C},]$.
Thus we have $\boldsymbol{g}_i = p(\MF^{(i)})$, where $p$ is a symmetrical pooling function.
The dynamic pooling on all group features $\MG$ of shape $[\texttt{M,C}]$ is formulated as $\MG = p(\MF, \MI)$.
The dynamic broadcast can be viewed as the inverse operation to dynamic pooling, which broadcasts $\boldsymbol{g}_i$ to all the points in the $i$-th group.
Since the broadcasting is essentially an indexing operation, we use the indexing notation $[\,]$ to denote it as $\MG[\MI]$, which is in the shape of $[\texttt{N,C}]$.
Dynamic broadcast/pooling is very efficient because it can be implemented with high parallelism on modern devices and well fits the sparse data with dynamic size. We provide an efficient implementation and runtime evaluation in Appendix~\ref{sec:appendix}.

The prerequisite of dynamic broadcast/pooling is that each point \emph{uniquely} belongs to a group, i.e. groups should not overlap with each other.
Thanks to the fact that there is no overlap among instances in the real 3D world, the groups do not overlap with each other naturally.

\subsubsection{Formulation of Sparse Instance Recognition}
After grouping points into instances in Sec.~\ref{sec:grouping}, we can directly extract instance features via some basic point-based networks like PointNet, DGCNN, etc.
There are three elements to define a basic point-based module: \textit{group center}, \textit{pair-wise feature} and \textit{group feature aggregation}.

\mypar{Group center}
The group center is the representative point of a group.
For example, in the ball query, it is the local origin of the sphere.
In SIR, the group center is defined as the centroid of all voted centers in a group.
\mypar{Pair-wise feature} defines the way to pair group center and neighbor points input for group-aware neighbor point feature extraction.
SIR adopts two kinds of pair-wise features: 1) the relative coordinate between the group center and each point, 2) the concatenation of the group feature and each point feature.
Taking feature concatenation as an example and using the notations in~\ref{subsec:preliminary}, the pair-wise feature can be denoted as $\texttt{CAT}(\MF,  \MG[\MI])$, where $\texttt{CAT}$ is channel concatenation.
\mypar{Group feature aggregation}
In a group, a pooling function is used to aggregate neighbor features.
SIR applies dynamic pooling to aggregate feature array $\MF$.
Following the notations in~\ref{subsec:preliminary}, we have $\MG = p(\MF, \MI)$, where $\MG$ is the aggregated group features.
\mypar{Integration}
Combining the three basic elements, we could build many variants of point-based operators, such as PointNet~\cite{pointnet}, DGCNN~\cite{dgcnn}, Meta-Kernel~\cite{rangedet}, etc.
Fig.~\ref{fig:dynamic_op} illustrates the basic idea of how to build an instance-level point operator with dynamic broadcast/pooling.
In our design, we adopt the formulation of VFE~\cite{voxelnet} as the basic structure of SIR layers, which is basically a two-layer PointNet.
In the $l$-th layer of SIR module, given the input point-wise feature array $\MF_l$, point coordinates array $\MX$, the voted center $\MX^\prime$ and group ID array $\MI$, the output of $l$-th layer can be formulated as:
\begin{align}
\label{eq:dir_layer1}
    \MF^{\prime}_l &= \texttt{LinNormAct}\left(\texttt{CAT} \left(\MF_l, \MX - p_{\text{avg}}(\MX^\prime, \MI)[\MI]\right) \right), \\
\label{eq:dir_layer2}
    \MF_{l+1} &= \texttt{LinNormAct}\left( \texttt{CAT}\left( \MF^{\prime}_l, p_{\text{max}}(\MF^{\prime}_l, \MI)[\MI] \right) \right),
\end{align}
where $\texttt{LinNormAct}$ is a fully-connected layer followed by a normalization layer~\cite{transformer} and an activation function~\cite{gelu}.
The $p_{\text{avg}}$ and the $p_{\text{max}}$ are average-pooling and max-pooling function, respectively.
The output $\MF_{l+1}$ can be further used as the input of the next SIR layer, so our SIR module is a stack of a couple of basic SIR layers.
\subsubsection{Sparse Prediction}
\label{sec:sparse_prediction}
With the formulation in Eqn.~\ref{eq:dir_layer1} and Eqn.~\ref{eq:dir_layer2}, SIR extracts features of all instances dynamically in parallel.
And then SIR makes \textbf{sparse prediction} for all groups. In contrast to two-stage sparse prediction, our proposals (i.e., groups) do not overlap with each other.
Unlike one-stage dense prediction, we only generate a single prediction for a group, which significantly reduces the cost of prediction head.
It is noteworthy that the fully sparse architecture may face a severe imbalance problem: short-range objects contain much more points than long-range objects.
Some methods~\cite{rcd, rangedet} use hand-crafted normalization factors to mitigate the imbalance.
Instead, SIR avoids the imbalance because it only generates a single prediction for a group regardless of the number of points in the group.
In most cases, a group corresponds to only a single ground truth box.

Specifically, for each SIR layer, there is a $\MG_l = p_{max}(\MF^{\prime}_l, \MI)$ in Eqn.~\ref{eq:dir_layer2}, which can be viewed as the group features.
We concatenate all $\MG_l$ from each SIR layer in channel dimension and use the concatenated group features to predict bounding boxes and class labels via MLPs.
All the groups whose centers fall into ground-truth boxes are positive samples.
For positive samples, the regression branch predicts the offsets from group centers to ground-truth centers and object sizes and orientations.
L1 loss~\cite{fasterrcnn} and Focal Loss~\cite{focalloss} are adopted as regression loss $\ML_{reg}$ and classification loss $\ML_{cls}$, respectively.

\subsection{Group Correction}
\label{sec:correction}
There is inevitable incorrect grouping in the Instance Point Grouping module.
For example, some foreground points may be missed, or some groups may be contaminated by background clutter.
So we leverage the bounding box proposals from SIR to correct the grouping.
The points inside a proposal belong to a corrected group regardless of their previous group IDs.
Since a few points may fall into multiple proposals, we simply make copies for these points along with their features and assign different copies to difference proposals.
After correction, we apply an additional SIR to these new groups.
To distinguish it from the first SIR module, we denote the additional SIR module as SIR2.

SIR2 predicts box residual from the proposal to its corresponding ground-truth box, following many two-stage detectors.
To make SIR2 aware of the size and location of a proposal, we adopt the offsets from inside points to proposal boundaries as extra point features following~\cite{lidarrcnn}. 
The regression loss is denoted as $\ML_{res} = \ML1(\Delta_{res}, \widehat{\Delta_{res}})$, where $\Delta_{res}$ is the ground-truth residual and $\widehat{\Delta_{res}}$ is the predicted residual.
Following previous methods~\cite{parta2, pvrcnn}, the 3D Intersection over Union (IoU) between the proposal and ground-truth serves as the soft classification label in SIR2.
Specifically, the soft label $q$ is defined as $q = \min(1, \max(0, 2IoU - 0.5))$, where $IoU$ is the area of Intersection over Union (IoU) between proposals and corresponding ground-truths.
Then cross entropy loss is adopted to train the classification branch, denoted as $\ML_{iou}$.
Taking all the loss functions in grouping (Sec.~\ref{sec:grouping}) and sparse prediction into account, we have
\begin{equation}
    \ML_{total} = \ML_{sem} + \ML_{vote} + \ML_{reg} + \ML_{cls} + \ML_{res} + \ML_{iou},
\end{equation}
where we omit the weight of each term for simplicity.
\subsection{Discussion}
\label{subsec:discussion}
The center voting in FSD is inspired by VoteNet~\cite{votenet}, while FSD has two essential differences from VoteNet.
\begin{itemize}[leftmargin=*]
    \item 
    After voting, VoteNet simply aggregates features around the voted centers without further feature extraction.
    FSD goes beyond this and builds a highly efficient SIR module taking advantage of dynamic broadcast/pooling, allowing for further instance-level feature extraction.
    Thus, FSD extracts more powerful instance features, which is experimentally demonstrated in Sec.~\ref{subsec:ablation}.
    \item
    VoteNet is a typical point-based method. As we discussed in Sec.~\ref{sec:intro}, it aggressively downsamples the whole scene to a fixed number of points for efficiency, causing inevitable information loss.
    Instead, the dynamic characteristic and efficiency of SIR enable fine-grained point feature extraction from any number of input points without any downsampling.
    In Sec.~\ref{subsec:ablation}, we showcase the efficiency of our design in processing large-scale point clouds and the benefits of fine-grained point representation.
\end{itemize}

\section{FSD++: FSD with Super Sparse Input}
It is well known that aggregating multiple frames as an input benefits performance. However, naive aggregation could result in a denser point cloud, which slows down the algorithm significantly, especially in the architecture with sparse operations.
This motivates us to pursue more sparse input data by removing temporal redundancy from the original point cloud stream.
Thanks to the fully sparse characteristic, the fully sparse model could greatly benefit from the increase of sparsity after redundancy removal.
Thus a natural question arises: \emph{How can we remove the redundancy while retaining the informative parts in advance?}
The similarities between consecutive point cloud frames offer us a potential solution to this question.
\par
In particular, the spatial distribution of points varies continuously and smoothly in a sequence.
We name the points that change between consecutive frames as \emph{residual points}.
The residual points are informative since they represent new observations in a time step.
Combining the residual points and history predictions, detectors have sufficient knowledge to infer about current objects.
In this paradigm, the residual points and previous foreground points together form a \emph{super sparse point cloud}.
FSD could directly take them as input for much more efficient object detection.
\par
\begin{figure*}[!t]
	\centering
	\includegraphics[width=0.98\linewidth]{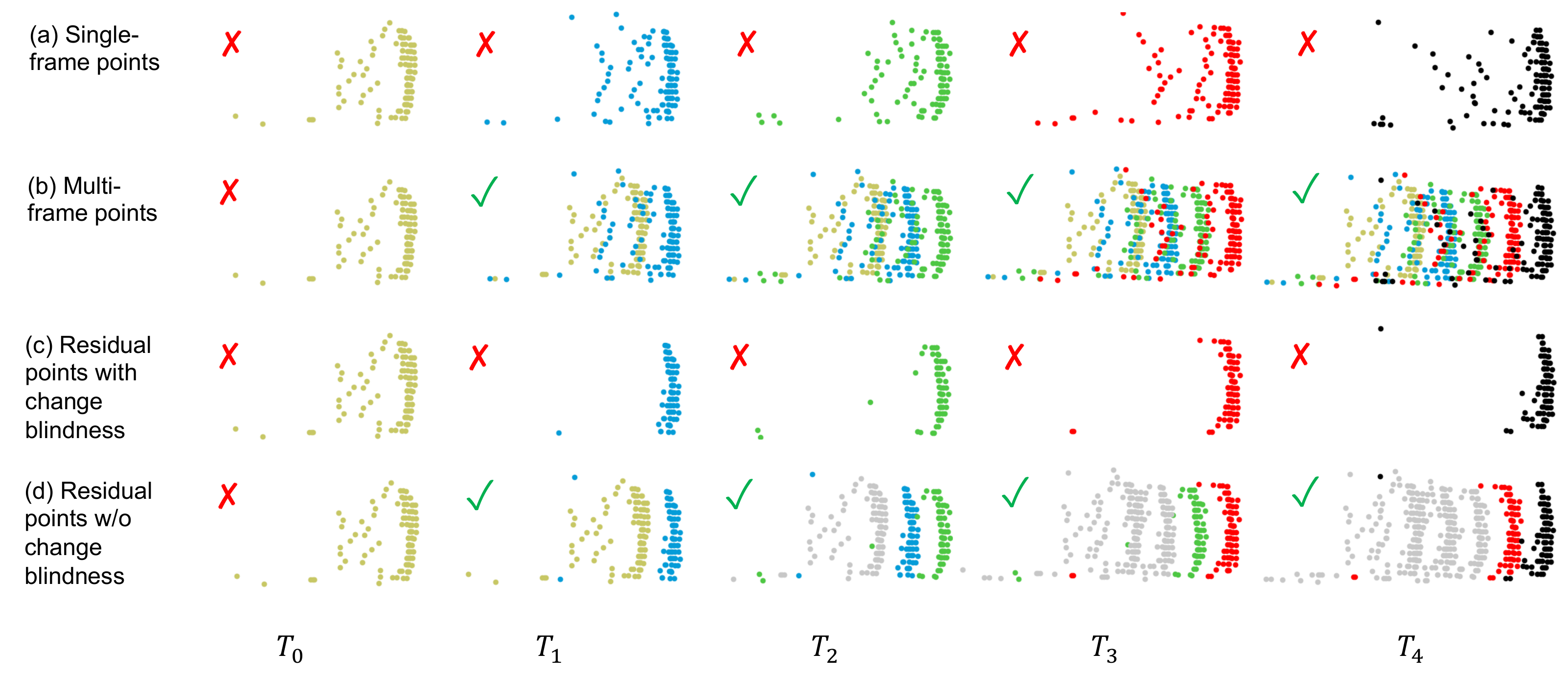}
	\caption{
            Illustration of temporal point manipulations.
            Different colors indicate points from different time steps.
            \gray{Gray} points are the sampled skeleton points.
            \red{Red cross \xmark} means the observation is too weak to be recognized as foreground objects (not really, just for illustration).
            \dgreen{Green check mark \cmark} means the observation is strong enough.
            (a) Points from a single frame are too weak.
            (b) After multi-frame point aggregation, the detector generates true positives from time step $T_1$.
            (c) Single-frame residual points suffer from change blindness. Since the detector cannot generate a true positive in $T_0$ for skeleton point sampling, we still only have the weak residual points in $T_1$. In this way, the detector outputs false negatives all the time.
            (d) Detector makes the right prediction in $T_1$ with residual points from two frames (max age is 2). So in the $T_2$, the detector could leverage the prediction in $T_1$ for skeleton point sampling, alleviating the change blindness.
        }
	\label{fig:change_blindness}
\end{figure*}
\subsection{Residual Points Probing}
\label{sec:res_points}
LiDAR sensors capture plenty of newly observed foreground points at each time step.
These points can be attributed to two main sources: (1) objects moving to new positions; (2) occluded regions becoming visible.
These newly observed points are referred to as residual points.
Residual points are critical to locate moving objects and detect recently emerged objects.
As we mentioned before, the residual points could be detected from the changes of point spatial distribution.

The residual point detection algorithm must fulfill two key requirements.
(1) The algorithm is supposed to be robust to tiny disturbances of points, which might be caused by sensing noise or tiny ego-motion estimation error.
It is unexpected that such point disturbances are detected as residual points.
(2) The algorithm should be highly efficient to handle millions of points from multiple frames.
In particular, each frame in WOD contains up to 200,000 points.

Several straightforward solutions meet the first requirement, i.e. ball query or voxelization into dense occupancy maps.
A point can be viewed as a residual point if no previous points fall into its neighborhood defined by the ball query radii.
The residual points can also be detected by the simple difference between the two dense occupancy maps.
Although proper ball query radii or voxel sizes bring robustness to point disturbances, these solutions still come with either high computational complexity ($O(N^2)$) or a huge memory footprint.

To fulfill both of the demands outlined above, we resort to hashing and design an algorithm shown in Algo.~\ref{alg:RPP}, named \emph{Residual Points Probing (RPP)}.
RPP consists of two steps.
(1) It first quantizes the point coordinates into integers. 
The granularity of quantization controls the robustness to point disturbances.
(2) For each point, RPP verifies if it is a residual point by hash probing.
Specifically, RPP first builds a hash table from previous quantized points.
The key set of the hash table is denoted as $K \subset \mathbb{Z}^{3}$, which is the quantized integer coordinates.
And value set of the hash table is denoted as $V = \{1, 0\}$, where $1$ indicates the slot is occupied and $0$ indicates the slot is unoccupied.
RPP then uses current quantized coordinates to probe the hash table.
If a current point hits an unoccupied slot, it is treated as a residual point.
Here is a hidden assumption in RPP that we assume two points are the same if they occupy the same voxel after quantization.
We adopt the well-known \emph{open addressing} for probing and the \emph{double hashing} as the hash function to reduce hash collisions.

\RestyleAlgo{ruled}
\SetAlgoNoLine
\SetKwComment{Comment}{/* }{ */}
\begin{algorithm}
\caption{Efficient Residual Points Probing}\label{alg:RPP}
\KwIn{current points $P_{cur}$, previous points $P_{pre}$, voxel size $s$, load factor $\alpha$,
hash function $h$}
\KwOut{Residual points of current frame $\Delta P_{cur}$}
\BlankLine
$\widehat{P}_{cur} \gets \texttt{Quantize}(P_{cur}, s)$\; $\widehat{P}_{pre} \gets \texttt{Quantize}(P_{pre}, s)$\;
Initialize empty hash table $T$ of length $|\widehat{P}_{pre}|/\alpha$\;
Initialize empty residual point set $\Delta P_{cur}$\;
\ForEach{$\widehat{p}_i$ in $\widehat{P}_{pre}$}{
$slot_i \gets \texttt{Probe}(T, h(\widehat{p}_i))$\;
\If{$slot_i$ is not occupied}{
$slot_i \gets \texttt{occupied flag}$\;
}
}
\ForEach{$\widehat{p}_i$ in $\widehat{P}_{cur}$}{
$slot_i \gets \texttt{Probe}(T, h(\widehat{p}_i))$\;
\If{$slot_i$ is not occupied}{
Add $p_i$ to $\Delta P_{cur}$\;
}
}
\Return{$\Delta P_{cur}$}
\end{algorithm}

Residual Point Probing is efficient in terms of both memory and speed.
For instance, we assume there are $N$ unique quantized coordinates in a point cloud clip.
If we expect the collision rate less than $\alpha$, the length of the hash table should be $N/\alpha$.
Empirically, $N$ is around 500,000 in a 5-frame point cloud in WOD.
A slot state can be represented as a single bit, and let $\alpha$ equal to 0.1.
We have that the memory cost of this hash table is around  0.6MB.
Moreover, the probing of each point is independent with each other, allowing for high parallelism in GPU.
\par
Formally, we denote the RPP process as follows:
\begin{equation}
    \Delta P_{t} = P_{t} - \bigcup_{i=1}^{B}P_{t-i},
    \label{eqn:rpp}
\end{equation}
where $\Delta P_{t}$ is the detected residual points in time step $t$ and $P_{t}$ is the raw points in time step $t$. The notation ``$X-Y$'' means removing the intersection of $X$ and $Y$ from $X$, equivalent to $X \setminus (X \cap Y)$. And the union means point cloud concatenation. $B$ is the number of previous frames used in RPP, which we term as \emph{base frames}.

\subsection{Skeleton Point Sampling}
\label{sec:skeleton_sampling}
Since residual points contain only new observations of the current time step, detectors require additional information from previous frames for sufficient input.
To incorporate this historical data, we use previously predicted boxes to crop previous foreground points, while discarding others outside of the boxes.
The cropped points are placed into the current frame after ego-motion compensation.
However, the foreground points from multiple previous frames are still essentially redundant.
Especially, the quite many points on short-range objects from multiple frames could lead to unnecessary overhead.
\par
To reduce the redundancy from multi-frame foreground points, we further sample within these cropped points.
Intuitively, we expect the sampled points contain the minimal information models need to make proper predictions.
In this sense, we refer to such a minimal subset of cropped points as \emph{skeleton points}, because they depict the basic structure or ``skeleton'' of objects.
Specifically, we try three kinds of sampling methods: \emph{random sampling}, \emph{farthest points sampling}, and \emph{voxel sampling}.
All the sampling methods are applied inside the previously predicted bounding boxes.
For random sampling and farthest points sampling, we adopt a predefined maximal point threshold $N_T$. We sample $N_T$ points inside the bounding boxes which contain points more than $N_T$.
For voxel sampling, we adopt dynamic voxelization~\cite{MVF} to voxelize points.
All the points falling into a voxel are reduced to a single point by average pooling.
\subsection{Treatment to Change Blindness}
\label{sec:change_blindness}
Theoretically, by combining the skeleton points and residual points, a model is able to make predictions in current frames.
However, a phenomenon known as ``change blindness'' can hinder performance.
Change blindness refers to the human visual system's tendency to overlook progressive small changes in a scene, even if the aggregated changes of multiple time steps are significant. 
A similar issue can occur in our case.
Thinking of a vehicle nearly entering into the sensing range of LiDARs in time step $t$, only a small part of the vehicle can be observed.
The detector is very likely to recognize it as background, so RPP will remove these points in time step $t+1$ and only keep a small number of new points of the vehicle as residual points.
In this way, if the vehicle appears slowly, the detector might never recognize it.
Fig. \ref{fig:change_blindness} demonstrates the change blindness.

To remedy the change blindness, we introduce a \emph{max age} $M$ for residual points.
In other words, the detector takes residual points from at most $M$ steps as input. Formally, the detector takes $\bigcup_{i=0}^{M-1}\Delta P_{t-i}$ 
as accumulated residual points for input in time step $t$.
\subsection{Integrated Super Sparse Input}
\begin{figure*}[h]
	\centering
	\includegraphics[width=0.98\linewidth]{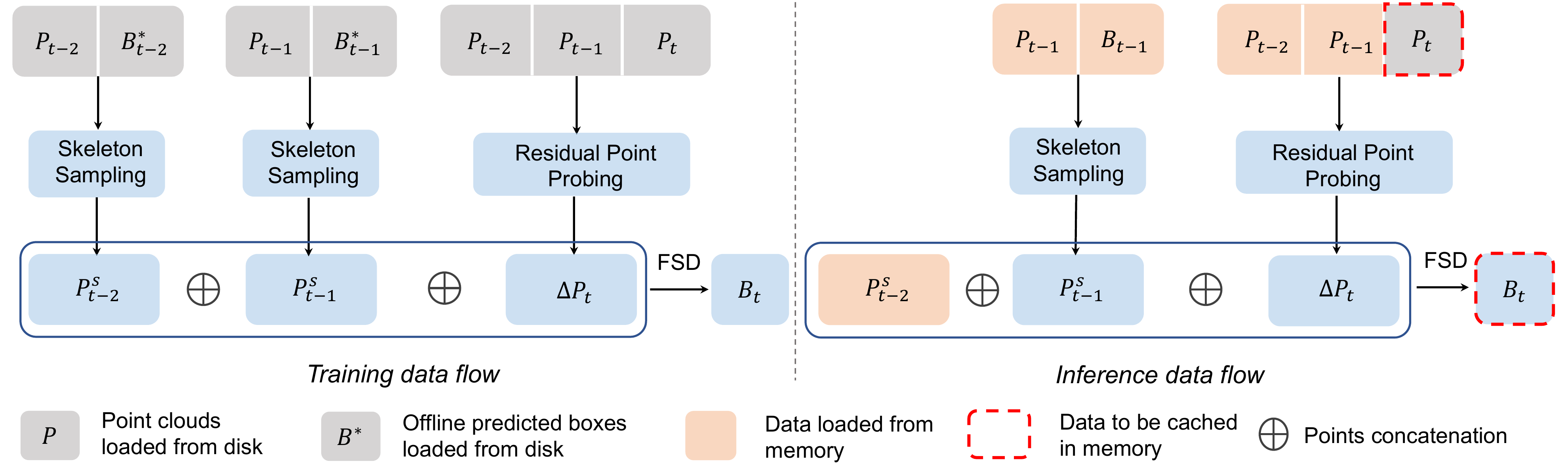}
	\caption{The overall architecture of FSD++. In training, we adopt offline predictions to approximate history predictions. During inference, the detector uses previous online predictions for skeleton point sampling. When the max age is larger than one, there will be some other $\Delta P_{t-i}$ from time step $t-i$. For simplicity, we only present $\Delta P_{t}$ here. }
	\label{fig:fsdpp_overall}
\end{figure*}
\begin{figure*}[h]
	\centering
	\includegraphics[width=0.98\linewidth]{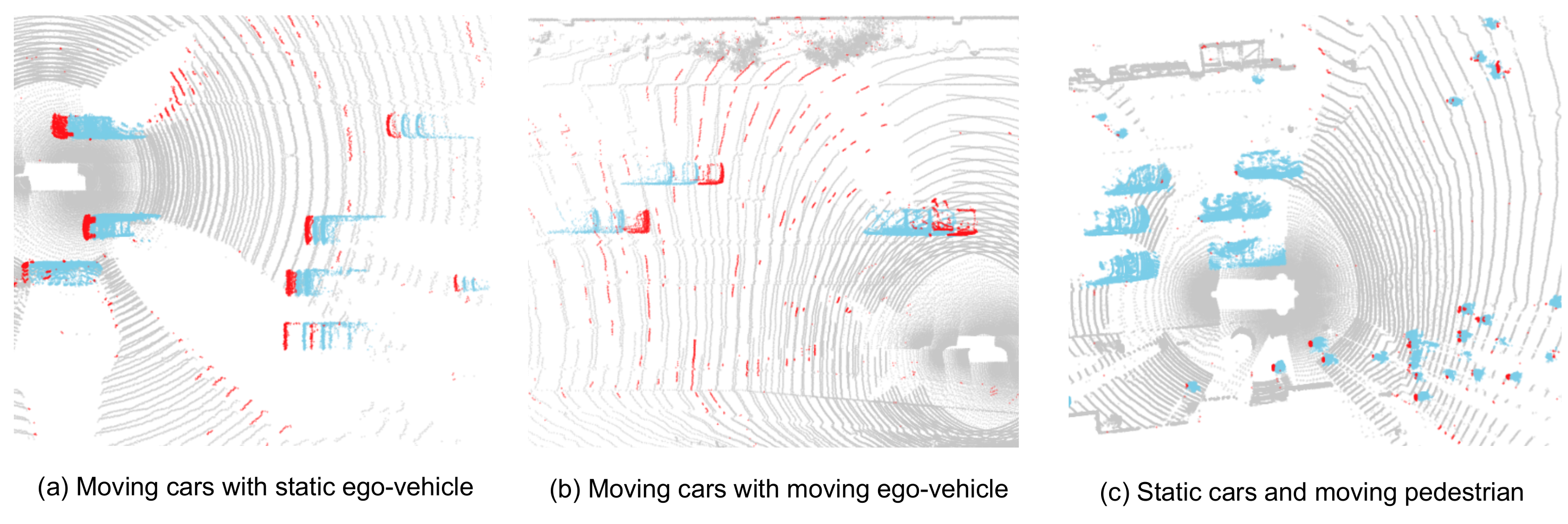}
	\caption{Examples of super sparse input point clouds.
        Residual points are colored in \red{red}.
        Previous foreground points are in \cyan{blue}.
        \gray{Gray} points will not be sent into the detector.
        (a) Moving cars cause apparent residual points. And some occluded points in the ground plane become visible due to car movement.
        (b) The ego-vehicle is moving, which causes some points in the ground plane to be detected as residual points.
        (c) Residual points are detected on the moving pedestrian instead of the static cars. 
        }
	\label{fig:input_points}
\end{figure*}

The input point clouds consist of two parts: previous skeleton points and residual points from multiple time steps.
Formally, for an $N$-frame FSD++ detector, we have the final input points in time step $t$ as follows:
\begin{equation}
    P_{t}^{in} = \left(\bigcup_{i=1}^{N}P^{s}_{t-i}\right) \cup \left(\bigcup_{i=0}^{M-1}\Delta P_{t-i}\right),
\end{equation}
where $P^{s}_t$ is the skeleton points at time step $t$.
$P^{in}_{t}$ is much more sparse than the raw point clouds and directly sent into the FSD detector.
Fig. \ref{fig:input_points} shows examples of $P^{s}$, $\Delta P$ and $P^{in}$.
\subsection{Training and Inference Pipeline}
The training and inference pipeline for FSD++ differs from the standard approach due to its use of history predictions and temporal information. Fig.~\ref{fig:fsdpp_overall} summarizes the overall pipeline. 
\subsubsection{Training}
To utilize history predictions, the input point cloud stream must be arranged in the temporal order.
However, the ordered input stream affects the model training due to a lack of data shuffling. 
Another problem is that the history predictions are not reliable in the early training stages.
Considering these two issues, we use a well-trained FSD detector to generate offline predictions of the entire training set. During training, for every sample, we load it along with its previous offline predictions to sample skeleton points.
Ground truth boxes seem to be an alternative to the offline predictions.
However, the distribution gap between ground truth used in training and predicted boxes used in inference is considerable.
Thus we adopt the offline predictions instead of ground-truth boxes.
\subsubsection{Inference}
\label{sec:fsdpp_inference}
In the online inference phase, the input point cloud stream is naturally in the temporal order.
For a point cloud sequence, the predictions of its first frame are from the well-trained FSD detector.
These predictions are regarded as the ``previous predictions'' of the first frame, which are called the \emph{seed predictions} of a sequence. 
We maintain several queues to cache some historical data that could be used more than once. For example, in a $N$-frame FSD++ pipeline, the raw points and skeleton points of time step $t$ could be reused from time step $t+1$ to $t+N-1$.

\section{Experiments}
\subsection{Setup}
\label{sec:setup}
\subsubsection{Dataset}
\noindent \textbf{Waymo Open Dataset (WOD)}
In our experiments, we use WOD~\cite{WOD} as the primary dataset to evaluate the performance of our proposed method.
WOD is the most trustworthy benchmark for LiDAR-based 3D object detection.
With 1150 sequences and more than 200,000 frames, WOD is currently the largest dataset of its kind.
Among them, 798 sequences are used for training, 202 for validation, and 150 for testing.
The detection range in WOD is 75 meters (cover area of $150m \times 150m$).

\noindent \textbf{Argoverse 2 (AV2)}
We further conduct long-range experiments on the recently released Argoverse 2 dataset~\cite{argo2} to demonstrate the superiority of FSD in long-range detection. 
AV2 has a similar scale to WOD, and it contains 1000 sequences in total, 700 for training, 150 for validation, and 150 for testing.
In addition to \textit{average precision} (AP), AV2 adopts a \textit{composite score} as an evaluation metric, which takes both AP and localization errors into account.
The perception range in AV2 is 200 meters (cover area of $400m \times 400m$), which is much larger than WOD. Such a large perception range leads to a huge memory footprint for dense detectors. 

\subsubsection{Model Configuration}
To demonstrate the generality of SIR, we build two FSD variants. FSD$_{sst}$ adopts the emerging single stride sparse transformer~\cite{sst} as the sparse voxel feature extractor. FSD$_{spconv}$ is built upon sparse convolution based U-Net in PartA2~\cite{parta2}.
In the experiments of FSD, we use FSD$_{sst}$ in the experiments unless otherwise specified.
In the experiments of FSD++, we use FSD$_{spconv}$ as our detector since the highly optimized engineering of SpConv makes it more efficient than the SST backbone with multi-frame input.

\subsubsection{Implementation Details}
Our implementation is based on popular MMDetection3D(v0.15)~\cite{mmdet3d}.
In FSD$_{sst}$, we use 4 sparse regional attention blocks~\cite{sst} as our voxel feature extractor.
The SIR module and SIR2 module consist of 3 and 6 SIR layers, respectively.
A SIR layer is defined by Eqn.~\ref{eq:dir_layer1} and Eqn.~\ref{eq:dir_layer2}.
Our SST-based model converges much faster than SST, so we train our models for 6 epochs for ablation study, instead of the $2\times$ schedule (24 epochs) in SST. 
For FSD$_{spconv}$, in addition to the 6-epoch schedule, we adopt a longer schedule (12 epochs) for better performance. Different from the default setting in MMDetection3D, we decrease the number of pasted instances in the CopyPaste augmentation.
In FSD, some scarce classes like cyclist prone to be over-fitted with too many pasted instances.
All experiments in Argoverse 2 dataset adopt a 12-epoch schedule.
The models for performance analysis (Sec.~\ref{subsec:inspect} $\sim$ Sec.~\ref{sec:fsdpp_analysis}) are trained on 8 RTX 2080Ti GPUs with batch-size 2. And the models in Table.~\ref{tab:sota} are trained on 8 RTX 3090 GPUs with batch-size 2. More details can be found in our released code.

\subsection{Main Results of FSD and FSD++}
We first compare FSD with state-of-the-art detectors and our baseline in Table \ref{tab:sota} and Table~\ref{tab:sota_test}.
In the validation split, FSD/FSD++ achieves state-of-the-art average performance (L2 mAPH) in single-frame/multi-frame settings, respectively.
In test split, FSD achieves the best performance on all classes among all single-frame detectors. Meanwhile, FSD++ with 7-frame input surpasses all detectors with up to 100-frame input, in terms of average metric.
\par
It is also noteworthy that FSD and FSD++ are much more efficient than most of the previous arts, especially in the multi-frame setting and long-range setting. We elaborate this in Sec.~\ref{sec:long_range} and  Sec.~\ref{sec:runtime}.
\begin{table*}[ht]
\small
\centering

\resizebox{0.99\linewidth}{!}{
\begin{tabular}{l|c|c|c|c|c|c|c|c}
  \specialrule{1pt}{0pt}{1pt}
\toprule
\multirow{2}{*}{Methods} & \multirow{2}{*}{\shortstack[1]{\#.\\ frames}} & \multirow{2}{*}{\shortstack[1]{mAP/mAPH\\ L2}} & \multicolumn{2}{c|}{\textit{Vehicle} 3D AP/APH} & \multicolumn{2}{c|}{\textit{Pedestrian} 3D AP/APH} & \multicolumn{2}{c}{\textit{Cyclist} 3D AP/APH}\\
 & &&   L1      &   L2     &   L1      &   L2  &   L1     &   L2\\
\midrule

SECOND~\cite{second} & 1 &61.0/57.2 & 72.3/71.7 & 63.9/63.3 & 68.7/58.2 & 60.7/51.3 & 60.6/59.3 & 58.3/57.0\\
MVF~\cite{MVF} & 1 &-/- & 62.9/-      & -/-   & 65.3/- & -/- & -/- & -/-\\

AFDet~\cite{afdet}& 1 & -/- & 63.7/-  & -/- & -/- & -/- & -/- & -/-\\
Pillar-OD~\cite{pillarbased} & 1 & -/- & 69.8/-   & -/-  & 72.5/-   & -/- & -/- & -/-\\

RangeDet~\cite{rangedet}&1 &65.0/63.2 & 72.9/72.3 & 64.0/63.6 & {75.9}/71.9 & 67.6/63.9 & 65.7/64.4 & 63.3/62.1 \\

PointPillars~\cite{pointpillar}&1 & 62.8/57.8 & 72.1/71.5  & 63.6/63.1 & 70.6/56.7  & 62.8/50.3 & 64.4/62.3 & 61.9/59.9 \\

Voxel RCNN~\cite{voxelrcnn}&1 & -/- & 75.6/-  & 66.6/- & -/- & -/- & -/- & -/- \\
RCD~\cite{rcd} & 1& -/- & 69.0/68.5  & -/- & -/- & -/- & -/- & -/-\\
VoTr-TSD~\cite{votr}&1 & -/- & 74.9/74.3 & 65.9/65.3 & -/- & -/- & -/- & -/-\\
LiDAR-RCNN~\cite{lidarrcnn}&1 & 65.8/61.3 & 76.0/75.5 & 68.3/67.9 & 71.2/58.7 & 63.1/51.7 & 68.6/66.9 & 66.1/64.4\\
Pyramid RCNN~\cite{pyramidrcnn} &1 & -/- & 76.3/75.7 & 67.2/66.7 & -/- & -/- & -/- & -/-\\
Voxel-to-Point~\cite{voxeltopoint}& 1& -/- & 77.2/- & 69.8/- & -/- & -/- & -/- & -/-\\
3D-MAN~\cite{3dman}& 16 & -/- & 74.5/74.0 & 67.6/67.1 & 71.7/67.7 & 62.6/59.0 & -/- & -/- \\
M3DETR~\cite{m3detr}& 1& 61.8/58.7 & 75.7/75.1 & 66.6/66.0 & 65.0/56.4 & 56.0/48.4 & 65.4/64.2 & 62.7/61.5\\
Part-A2-Net~\cite{parta2}&1 & 66.9/63.8 & 77.1/76.5 & 68.5/68.0 & 75.2/66.9 & 66.2/58.6 & 68.6/67.4 & 66.1/64.9\\
CenterPoint-Pillar~\cite{centerpoint}& 1& -/- & 76.1/75.5 & 68.0/67.5 & 76.1/65.1 & 68.1/57.9 & -/- & -/-\\
CenterPoint-Voxel~\cite{centerpoint}&1 & 69.8/67.6 & 76.6/76.0 & 68.9/68.4 & 79.0/73.4 & {71.0}/65.8 & 72.1/71.0 & 69.5/68.5\\
IA-SSD~\cite{iassd}&1 & 62.3/58.1 & 70.5/69.7 & 61.6/61.0 & 69.4/58.5 & 60.3/50.7 & 67.7/65.3 & 65.0/62.7 \\

PV-RCNN~\cite{pvrcnn} & 1 &66.8/63.3& 77.5/76.9     & 69.0/68.4 & 75.0/65.6  &  66.0/57.6 & 67.8/66.4 & 65.4/64.0\\
RSN ~\cite{rsn}& 1& -/- & 75.1/74.6 &66.0/65.5 & 77.8/72.7 & 68.3/63.7 & -/- & -/- \\
SST\_TS~\cite{sst}&1 & -/- & 76.2/75.8 & 68.0/67.6 & 81.4/74.0 & 72.8/{65.9} & -/- & -/-\\
SST~\cite{sst}&1 & 67.8/64.6 & {74.2/73.8} & {65.5/65.1} & 78.7/69.6 & 70.0/61.7 & 70.7/69.6 & 68.0/66.9\\
AFDetV2~\cite{afdetv2}& 1& 71.0/68.8 & {77.6/77.1} & {69.7}/69.2 & 80.2/74.6 & 72.2/67.0 & 73.7/72.7 & 71.0/70.1\\
PillarNet-34~\cite{pillarnet}&1 & 71.0/68.5 & {79.1/78.6} & \bests{70.9}/\bests{70.5} & 80.6/74.0 & 72.3/66.2 & 72.3/71.2 & 69.7/68.7\\
PV-RCNN++~\cite{pvrcnnpp}& 1 & 68.4/64.9 &  {78.8}/{78.2}     &  {70.3}/{69.7}&  76.7/67.2  &  68.5/59.7 & 69.0/67.6 & 66.5/65.2\\
PV-RCNN++(center)~\cite{pvrcnnpp}&1  & 71.7/69.5 &  \bests{79.3}/\bests{78.8}   &  {70.6}/{70.2}&  81.3/76.3  &  73.2/68.0 & 73.7/72.7 & 71.2/70.2\\
CenterFormer~\cite{centerformer} & 8 & 75.1/73.7 & 78.8/78.3 & 74.3/73.8 & 82.1/79.3 & 77.8/75.0 & 75.2/74.4 & 73.2/72.3\\
INT ~\cite{int} & 10 & -/73.6 & -/- & -/73.3 & -/- & -/71.9 & -/- & -/75.6 \\
MPPNet~\cite{mppnet} & 16 & 75.6/74.9 & \best{82.7}/\best{82.3} & \best{75.4}/\best{75.0} & 84.7/82.3 & 77.4/75.1 & 77.3/76.7 & 75.1/74.5\\
\midrule
FSD$_{spconv}$ (ours) & 1& 71.9/69.7 &77.8/77.3 & 68.9/68.5 & 81.9/76.4 & 73.2/68.0 & 76.5/75.2 & 73.8/72.5 \\
FSD$_{sst}$ (ours) &1 & 71.5/69.2 & 76.8/76.3 & 67.9/67.5 & 81.3/75.3 & 72.5/67.0 & {77.2/76.0} & {74.4}/73.2 \\
FSD$_{spconv}$ (ours)\dag &1   & \bests{72.9}/\bests{70.8} & 79.2/{78.8} & 70.5/{70.1} & \bests{82.6}/\bests{77.3} & \bests{73.9}/\bests{69.1} & \bests{77.1}/\bests{76.0} & \bests{74.4}/\bests{73.3} \\
FSD++ (ours)\dag & 7 & \best{76.8}/\best{75.5} & 81.4/80.9 & 73.3/72.9 & \best{85.1}/\best{82.2} & \best{78.2}/\best{75.4} & \best{81.2}/\best{80.3} & \best{78.9}/\best{78.1} \\
\bottomrule
  \specialrule{1pt}{1pt}{0pt}
\end{tabular}
}
\vspace{2mm}
\caption{
    {Performances on the Waymo Open Dataset validation split.}
    All reported results are from single model without any test-time augmentations.
    \dag: Longer schedule (12 epochs).
    We mark the best single-frame results and multi-frame results with \bests{gray} boxes and \best{cyan} boxes, respectively.
    }
    \vspace{3mm}
 \label{tab:sota}

\end{table*}

\begin{table*}[ht]
\small
\centering

\resizebox{0.99\linewidth}{!}{
\begin{tabular}{l|c|c|c|c|c|c|c|c}
  \specialrule{1pt}{0pt}{1pt}
\toprule
\multirow{2}{*}{Methods} & \multirow{2}{*}{\shortstack[1]{\#.\\ frames}} & \multirow{2}{*}{\shortstack[1]{mAP/mAPH\\ L2}} & \multicolumn{2}{c|}{\textit{Vehicle} 3D AP/APH} & \multicolumn{2}{c|}{\textit{Pedestrian} 3D AP/APH} & \multicolumn{2}{c}{\textit{Cyclist} 3D AP/APH}\\
 & &&   L1      &   L2     &   L1      &   L2  &   L1     &   L2\\
\midrule

CenterPoint~\cite{centerpoint}& 1 & -/69.0 & -/- & -/71.9 & -/- & -/67.0 & -/- & -/68.2\\
AFDetV2-lite~\cite{afdetv2}& 1 & 72.2/70.0 & 80.5/80.0 & 73.0/72.6 & 79.8/74.3 & 73.7/68.6 & 72.4/71.2 & 69.8/69.7 \\
PV-RCNN~\cite{pvrcnn}& 1  &71.3/68.8 & 80.6/80.1     & 72.8/72.4 & 78.2/72.0  &  71.8/66.0 & 71.8/70.4 & 69.1/67.8\\
PV-RCNN++~\cite{pvrcnnpp}& 1  & 72.4/70.2 & 81.6/81.2    & 73.9/73.5 &  80.4/75.0  &  74.1/69.0 & 71.9/70.8 & 69.3/68.2\\
Graph R-CNN~\cite{graphrcnn} & 1  & 73.8/71.6 & \bests{83.6}/\bests{83.1}    & \bests{76.0}/\bests{75.6} &  81.9/76.5  &  75.6/70.5 & 72.5/71.3 & 69.8/68.7\\
AFDetV2~\cite{afdetv2}& 2 & 74.6/73.1 & 81.7/81.2 & 74.3/73.9 & 81.3/78.0 & 75.5/72.4 & 76.4/75.4 & 74.1/73.0  \\
CenterPoint++~\cite{centerpoint}& 3 & 74.2/72.8 & 82.8/82.3 & 75.5/75.1 & 81.0/78.2 & 75.1/72.4 & 74.4/73.3 & 72.0/71.0\\
BEVFusion$^\ast$~\cite{bevfusion}& 3 & 77.7/76.3 & \best{85.0}/\best{84.6} & 77.9/77.5 & \best{84.7}/\best{82.0} & 79.1/76.4 & 78.5/77.5 & 76.0/75.1\\
DeepFusion$^\ast$~\cite{deepfusion}& 5 & 76.9/75.5 & 83.3/82.8 & 76.1/75.6 & {84.6}/{81.8} & 79.2/76.4 & 77.8/76.8 & 75.5/74.5\\
CenterFormer~\cite{centerformer}& 16 & 76.9/75.6 & {84.7}/{84.4} & \best{78.1}/\best{77.7} &  {84.6}/{81.8} & \best{79.4}/\best{76.6} & 75.5/74.5 & 73.3/72.4\\
INT~\cite{int}& 100 & 76.6/75.2 & 84.7/84.3    & 78.0/77.6 &  82.4/79.7  &  76.6/74.0 & 77.4/76.3 & 75.2/74.1\\
MPPNet~\cite{mppnet}& 16 & 76.9/75.7 & 84.3/83.9 & 77.3/76.9 &  84.1/81.5 & 78.4/75.9 & 77.1/76.4 & 74.9/74.2\\
\midrule
FSD$_{spconv}$ (ours)\dag & 1 & \bests{74.4}/\bests{72.4} & {82.7}/{82.3} & {74.4}/{74.1} & \bests{82.9}/\bests{77.9} & \bests{75.9}/\bests{71.3} & \bests{75.6}/\bests{74.4} & \bests{72.9}/\bests{71.8} \\
FSD++ (ours)\dag & 7  & \best{78.4}/\best{77.1} & 84.5/84.1 & 77.1/76.7 & 84.5/81.7 & 79.0/76.2 & \best{81.4}/\best{80.5} & \best{79.2}/\best{78.3} \\

\bottomrule
  \specialrule{1pt}{1pt}{0pt}
\end{tabular}
}
\vspace{2mm}
\caption{
    {Performances on the Waymo Open Dataset test split. All results are in single-model setting without ensemble or test-time augmentations. $^\ast$: Multi-modal methods with camera information. We mark the best single-frame results and multi-frame results with \bests{gray} boxes and \best{cyan} boxes, respectively. \dag: 12-epoch schedule.
    }
    }
    \vspace{3mm}
 \label{tab:sota_test}

\end{table*}

\subsection{Study of Treatments to Center Feature Missing}
\label{subsec:inspect}
\subsubsection{Quantitative Experiments}
In what follows, we conduct experiments on WOD to investigate the issue of \textbf{Center Feature Missing (CFM)}.
We first develop several models with different characteristics.
Note that all the following models adopt the same voxelization resolution, so they face the same degree of CFM at the beginning.
\begin{itemize}[leftmargin=*]
    \item
    \textbf{FSD}$_{plain}$:
    After the sparse voxel encoder, FSD$_{plain}$ directly predicts the box from each voxel.
    The voxels inside ground-truth boxes are assigned as \textit{positive}.
    Although FSD$_{plain}$ uses the most straightforward solution for CFM, it suffers from the large variance of regression targets and low-quality predictions from informative voxels.
    \item
    \textbf{SST}$_{center}$:
    It replaces the anchor-based head in SST with  CenterHead~\cite{centernet, centerpoint}.
    Based on the sparse voxel encoder, SST$_{center}$ converts sparse voxels into dense feature maps and applies several convolutions to diffuse features to the empty object centers as in Fig.~\ref{fig:diffusion}.
    Then it makes predictions from the diffused center feature.
    \item
    \textbf{FSD}$_{nogc}$:
    It removes the group correction and SIR2 module in FSD.
    \item
    \textbf{CenterPoint-PP}:
    It does not resort to any sparse voxel encoders. Instead, it applies multiple dense convolutions soon after voxelization for feature diffusion, greatly eliminating CFM.
    It is also equipped with CenterHead to avoid large variance of regression targets.
    
\end{itemize}

\begin{table}

\small

\resizebox{0.99\columnwidth}{!}{
\begin{tabular}{l|cccc|c}
\toprule
 & \multicolumn{4}{c|}{Vehicle length (m)} & \\
Methods & [0, 4) & [4, 8) & [8, 12) & [12, $+\infty$) & Official$^\ast$\\

\midrule
CenterPoint-PP\dag                    & 34.3 & 69.3 & 42.0 & 43.6 & 66.2 \\
FSD$_\text{plain}$   & 32.2 & 64.6 & 41.3 & 42.2 & 62.3 \\
SST$_\text{center}$~\cite{sst}   & 36.0 & 69.4 & 33.7 & 30.5 & 66.3 \\
\midrule
FSD$_\text{nogc}$   & 33.5 \down{2.5} & 68.2 \down{1.2} & 47.7 \up{14.0} & 47.9 \up{17.4} & 65.2 \down{1.1} \\
FSD & 36.7 \up{0.7} & 71.0 \up{1.6} & 51.3 \up{17.6} & 53.7 \up{23.2} & 69.3 \up{3.0}   \\

\bottomrule
\end{tabular}
}
\vspace{2mm}
\caption{
    {Vehicle detection with vehicle length breakdown.}
    \dag: re-implemented ourselves.
    $^\ast$: official Waymo L2 overall metric.
    Arrows indicate the performance changes from SST$_{center}$.
    }
    \label{tab:large_vehicle}

\end{table}

There is usually a quite large unoccupied area around the centers of large vehicles.
Thus the performance of large vehicles is an appropriate indicator that reveals the effect of CFM.
So we build a customized evaluation tool, which breaks down the object length following the COCO evaluation~\cite{coco}.
Then we use it to evaluate the performance of vehicles with different lengths.
Table \ref{tab:large_vehicle} shows the results, and we list our findings as follows.
\begin{itemize}[leftmargin=*]
    \item
    Comparing FSD$_{plain}$ with SST$_{center}$, they share the same attention-based sparse voxel encoder. However, the trend is totally opposite w.r.t vehicle size.
    With feature diffusion, SST$_{center}$ attains much worse performance than FSD$_{plain}$ on large vehicles.
    It suggests feature diffusion is a sub-optimal solution for CFM in the case of large objects.
    For those large objects, the features may not be diffused to the centers or the diffused features are too weak to make accurate predictions.
    \item
    However, FSD$_{plain}$ obtains the worst performance among all detectors on vehicles with normal sizes.
    Note that the CFM issue is minor for the normal-size vehicles.
    So, in this case, the center-based assignment in SST$_{center}$ shows its superiority to the assignment in FSD$_{plain}$.
    It suggests the solution for CFM in FSD$_{plain}$ is also sub-optimal, even if it achieves better performance in large objects.
    \item
    Comparing FSD$_{nogc}$ with SST$_{center}$, they share the same sparse voxel encoder while FSD$_{nogc}$ replaces the dense part in SST$_{center}$ with SIR.
    The huge improvements of FSD$_{nogc}$ on large vehicles fairly reveal that SIR effectively resolves CFM and is better than feature diffusion.
    \item
    CenterPoint-PP suffers much less from CFM because it leverages dense feature maps from the very beginning of the network.
    It is also equipped with advanced center-based assignment.
    Even so, FSD$_{nogc}$ and FSD still outperform CenterPoint-PP, especially on large vehicles.
\end{itemize}
\subsubsection{Qualitative Analysis}
\begin{figure}[h]
	\centering
	\includegraphics[width=\columnwidth]{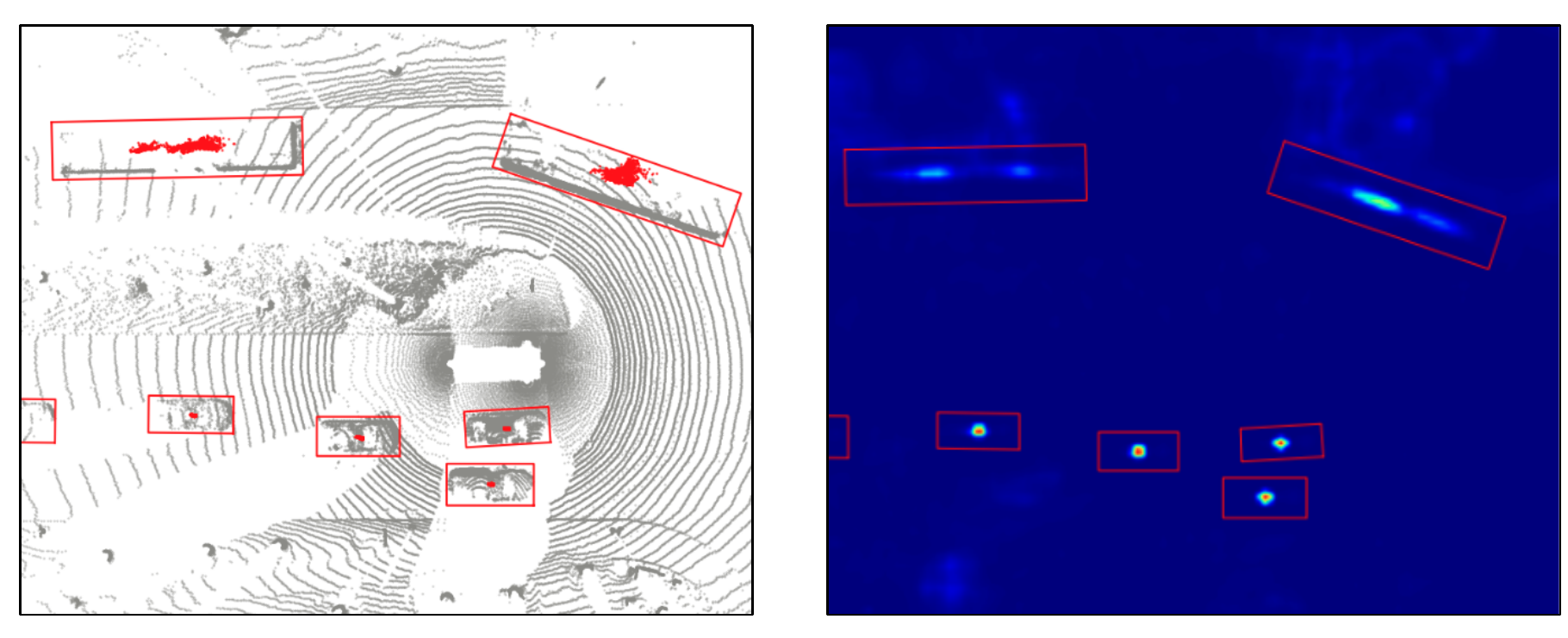}
	\vspace{-2mm}
	\caption{An intuitive illustration of the center feature missing. \textbf{Left}: Voted centers of FSD. \textbf{Right:} Predicted heatmap of SST$_{center}$. }
	\label{fig:cfm_vis}
\end{figure}

In addition to the quantitative experiments, we demonstrate the qualitative effect of CFM and our treatment, shown in Fig.~\ref{fig:cfm_vis}. 
Fig.~\ref{fig:cfm_vis} showcases the voted centers of FSD and the predicted heatmap of center-based SST.
Both of them yield high-quality predictions for vehicles of normal size, but their predictions (votes) are usually ambiguous for large vehicles.
Center-based dense detectors make predictions from such ambiguous heatmaps, so they are prone to make flawed final predictions.
Although the center voting of FSD on large vehicles is also mediocre, FSD only uses the votes to obtain point groups (i.e., instance segmentation), which does not necessitate perfect center voting.
The final predictions of FSD are derived from the complete point groups rather than the weak center features, thereby sidestepping the issue of CFM.

\begin{figure}[h]
	\centering
	\includegraphics[width=0.99\linewidth]{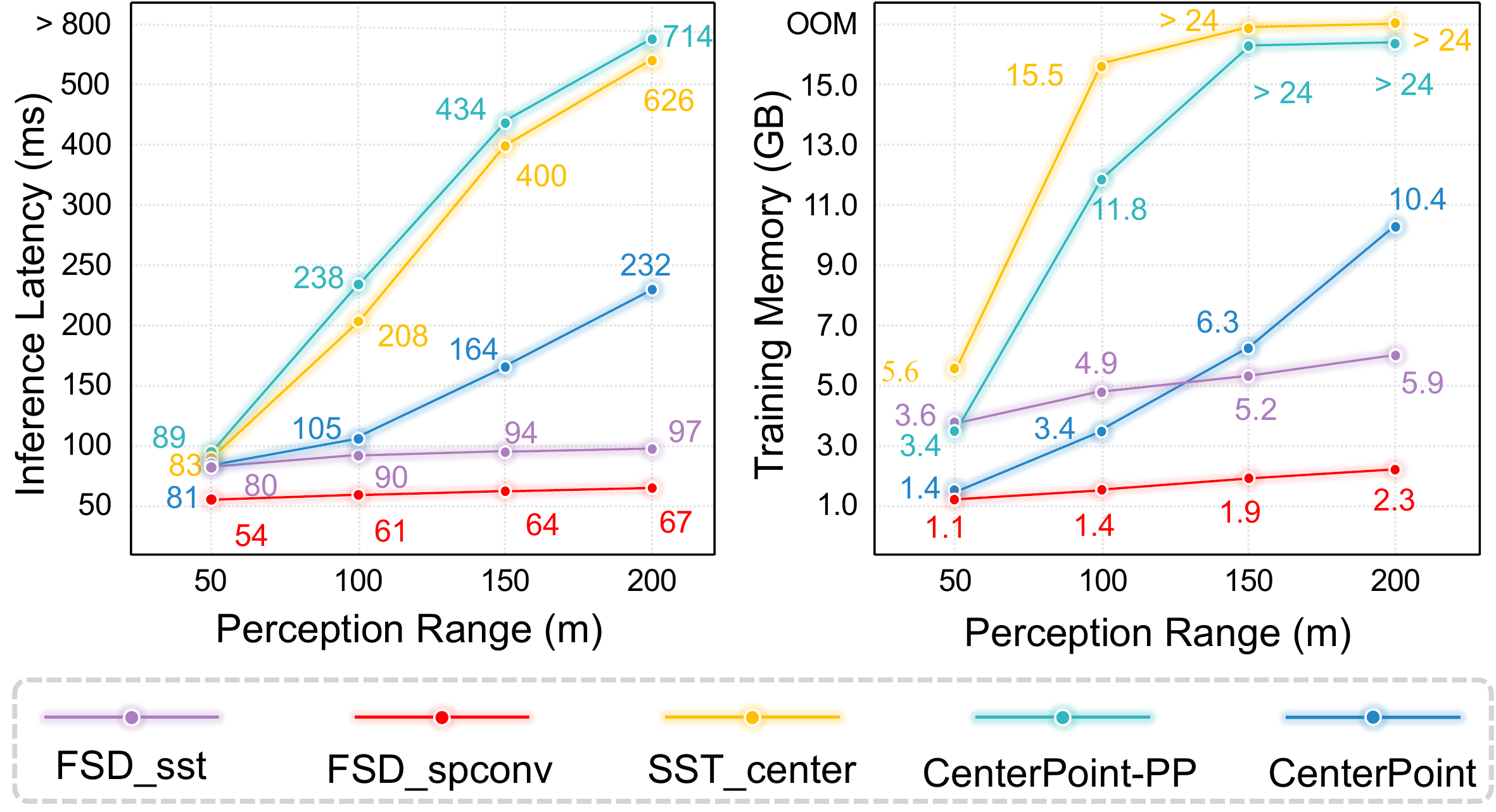}
	\caption{
	{Memory footprints and inference latency in different perception ranges.}
	Statistics are obtained on a single 3090 GPU with batch size 1.
	Inference latency is evaluated by the standard benchmark script in MMDetection3D without any test-time optimization.
	CenterPoint-PP and SST$_{center}$ are defined in Sec.~\ref{subsec:inspect}.
	Best viewed in color.
	}

	\label{fig:diff_range}
\end{figure}
\begin{table*}[ht]

\begin{center}
\resizebox{\textwidth}{!}{%
\begin{tabular}{l|ccccccccccccccccccccccccc}
\toprule
  \textbf{Methods} &
 \rotatebox{90} {\textbf{\em Average}} & 
 \rotatebox{90} {Vehicle} & 
 \rotatebox{90} {Bus} &
 \rotatebox{90} {\textbf{\em Pedestrian}} &
 \rotatebox{90} {\textbf{\em Stop Sign}} &
 \rotatebox{90} {Box Truck} &
 \rotatebox{90} {Bollard} &
 \rotatebox{90} {\textbf{\em C-Barrel}} &
 \rotatebox{90} {\textbf{\em Motorcyclist}} &
 \rotatebox{90} {MPC-Sign} &
 \rotatebox{90} {\textbf{\em Motorcycle}} &
 \rotatebox{90} {Bicycle} &
 \rotatebox{90} {\textbf{\em A-Bus}} &
 \rotatebox{90} {\textbf{\em School Bus}} &
 \rotatebox{90} {Truck Cab} &
 \rotatebox{90} {\textbf{\em C-Cone}} &
 \rotatebox{90} {V-Trailer} &
 \rotatebox{90} {Sign} &
 \rotatebox{90} {Large Vehicle} &
 \rotatebox{90} {\textbf{\em Stroller}} &
 \rotatebox{90} {\textbf{\em Bicyclist}} &
\\ 
 \midrule
\textit{Precision} &&&& \\
\midrule
CenterPoint\dag~\cite{centerpoint}        & 13.5 & 61.0 & 36.0  & 33.0  & 28.0  & 26.0  & 25.0  & 22.5  & 16.0   & 16.0 & 12.5 & 9.5  & 8.5 & 7.5 & 8.0 & 8.0  & 7.0 & 6.5  &3.0 & 2.0 & 14  \\
CenterPoint$^\ast$       & 22.0 & 67.6 & 38.9 & 46.5 & 16.9 & 37.4 & 40.1 & 32.2 & 28.6 & 27.4 & 33.4 & 24.5 & 8.7 & 25.8 & 22.6 & 29.5  & 22.4 & 6.3 & 3.9 & 0.5 & 20.1\\
FSD       & 24.0 & 67.1 & 39.8 & 57.4 & 21.3 & 38.3 & 38.3 & 38.1 & 30.0 & 23.6 & 38.1 & 25.5 & 15.6 & 30.0 & 20.1 & 38.9  & 23.9 & 7.9 & 5.1 & 5.7 & 27.0\\
FSD\ddag      & 28.2 & 68.1 & 40.9 & 59.0 & 29.0 & 38.5 & 41.8 & 42.6 & 39.7 & 26.2 & 49.0 & 38.6 & 20.4 & 30.5 & 14.8 & 41.2  & 26.9 & 11.9 & 5.9 & 13.8 & 33.4\\
\midrule
\textit{Composite Score} &&&& \\
\midrule
CenterPoint$^\ast$        & 17.6  & 57.2  & 32.0  & 35.7  & 13.2  & 31.0  & 28.9  & 25.6  & 22.2 & 19.1 & 28.2  & 19.6 & 6.8  & 22.5 & 17.4  & 22.4 & 17.2 & 4.8 & 3.0 & 0.4 & 16.7  \\
FSD                & 19.1 & 56.0   & 33.0  & 45.7  & 16.7  & 31.6  & 27.7  & 30.4  & 23.8 & 16.4 & 31.9  & 20.5 & 12.0 & 25.6 & 15.9  & 29.2 & 18.1 & 6.4 & 3.8 & 4.5 & 22.1 \\
FSD\ddag               & 22.7 & 57.7   & 34.2  & 47.5  & 23.4 & 31.7  & 30.9  & 34.4  & 32.3 & 18.0 & 41.4  & 32.0 & 15.9 & 26.1 & 11.0  & 30.7 & 20.5 & 9.5 & 4.4 & 11.5 & 28.0 \\
\bottomrule

\end{tabular}%
}
\end{center}
\caption{
{Performance in Argoverse 2 validation split.}
\dag: provided by authors of AV2 dataset.
\ddag: Weak CopyPaste augmentation for preventing overfitting (one instance per class).
$\ast$: re-implemented by ourselves.
C-Barrel: construction barrel.
MPC-Sign: mobile pedestrian crossing sign.
A-Bus: articulated bus.
C-Cone: construction cone.
V-Trailer: vehicular trailer.
We omit the results of dog, wheelchair and message board trailer because these categories contain very few instances.
The average results take all categories into account, including the omitted categories.
We mark the categories attaining notable improvements in \textbf{\em bold}.
}

\label{tab:argo_main}
\end{table*}

\subsection{Long-range Detection}
\label{sec:long_range}
Several widely adopted 3D detection benchmarks~\cite{WOD, kitti, nus} have relatively short perception range.
To unleash the potential of FSD, we conduct long-range detection experiments on the recently released Argoverse 2 dataset (AV2).
AV2 has a perception range up to 200 meters, making it an ideal testbed for our method.
In addition, AV2 contains objects in 30 classes, exhibiting the long-tail distribution, which is also another challenge for FSD.

\subsubsection{Main results}
We first list the main results of FSD on AV2 in Table \ref{tab:argo_main}.
The authors of AV2 provide a baseline CenterPoint model, but the results are mediocre.
To make a fair comparison, we re-implement a stronger CenterPoint model on the AV2 dataset.
The re-implemented CenterPoint adopts the same training scheme with FSD, including ground-truth sampling to alleviate the long-tail issue.
FSD outperforms CenterPoint in the average metric.
It is noteworthy that FSD significantly outperforms CenterPoint in some tiny objects (e.g., Pedestrian, Construction Cone) as well as some objects with extremely large sizes (e.g., Articulated Bus, School Bus).
We owe this to the virtue of instance-level fine-grained feature extraction in SIR.

\subsubsection{Range Scaling}
To demonstrate the efficiency of FSD in long-range detection, we depict the trend of training memory and inference latency of three detectors when the perception range increases in Fig.~\ref{fig:diff_range}.
Fig.~\ref{fig:diff_range} shows that dense detectors experience a dramatic increase in latency and memory footprint as the perception range grows.
Designed to be fully sparse, the resource needed for FSD is roughly linear to the number of input points, so its memory and latency only slightly increase as the perception range extends.

\subsection{Performance Inspection of FSD}
\label{subsec:ablation}
\subsubsection{Effectiveness of Components}
In addition to FSD$_{plain}$ and FSD$_{nogc}$ (Sec.~\ref{subsec:inspect}), we also degrade FSD to FSD$_{agg}$ to gain insights into its mechanism.
In FSD$_{agg}$, we aggregate grouped point features by dynamic pooling after Instance Point Grouping, and then directly make predictions from the pooled features.
FSD$_{agg}$ is similar to the way in VoteNet~\cite{votenet} as we discussed in Sec.~\ref{subsec:discussion}.
Thus, FSD$_{agg}$ can explicitly leverage instance-level features other than the point-level features in FSD$_{plain}$, mitigating the issue of CFM. 
However, FSD$_{agg}$ cannot take advantage of further point feature extraction in SIR.
As can be seen in Table~\ref{tab:ablation}, the improvement is limited if we only apply grouping without SIR.
The combination of grouping and SIR, on the other hand, yields significant improvements.
\begin{table}[H]
	\footnotesize 

	\begin{center}
		\resizebox{0.99\columnwidth}{!}{
			\begin{tabular}{l|ccc|ccc}
				\toprule
				\multirow{2}{*}{} &\multirow{2}{*}{Grouping} & \multirow{2}{*}{SIR} & \multirow{2}{*}{\shortstack[1]{Group\\Correction}} & 			
				\multicolumn{3}{c}{L2 3D APH}  \\
				&&&&Vehicle & Pedestrian & Cyclist\\
				\midrule
				$\text{FSD}_{plain}$ &   &  &  & 62.29 & 64.31 & 64.49\\
				$\text{FSD}_{agg}$ &\checkmark  &  &  & 63.13 & 65.13 & 64.52\\
				$\text{FSD}_{nogc}$ &\checkmark & \checkmark    &  & 65.20 & 67.39 & 67.78\\
				FSD &\checkmark & \checkmark & \checkmark  & 69.30 & 69.30 & 69.60\\

				 \bottomrule
			\end{tabular}
		}	
	\end{center}
		\caption{{Ablation of design factors in SIR.} Performances are evaluated on Waymo validation split.}
	\label{tab:ablation}

\end{table}

\subsubsection{Downsampling in SIR}
The efficiency of SIR makes it feasible to extract fine-grained point features without any point downsampling.
This is another notable difference between FSD and VoteNet.
To demonstrate the superiority, we apply voxelization on the raw points before the SIR module and treat the centroids of voxels as downsampled points.
We conduct experiments on the AV2 dataset because it contains a couple of categories in a tiny size, which may be sensitive to downsampling.
As expected, small objects have notable performance loss when adopting downsampling, and we list some of them in Table ~\ref{tab:argo_voxel}.
We also evaluate the inference latency of the SIR module on a 3090 GPU, which is highly efficient.
\begin{table}[h]
\small
\centering

\resizebox{0.99\columnwidth}{!}{
\begin{tabular}{l|cccc|c}
\toprule
& \multicolumn{4}{c|}{AP} \\
Voxel size & CC & Bollard  & Bicyclist & Stop Sign & Latency (ms)\dag\\

\midrule

30cm   & 35.4 & 36.5  & 24.6 & 18.3 & 3.5\\
20cm   & 37.3 & 37.3  & 26.4 & 20.0 & 4.1\\
10cm   & 38.9 & 38.3  & 27.0 & 21.3 & 4.5\\
Point  & 39.3 & 38.6  & 27.1 & 21.5 & 6.3\\

\bottomrule
\end{tabular}
}
\vspace{3mm}
\caption{
{Performances with different representation granularity.}
\dag: Latency of SIR module.
    }
    \label{tab:argo_voxel}
\end{table}

\subsubsection{HD Map-assisted Detection}
\begin{table}[h]

	\footnotesize 
	\begin{center}
		\resizebox{0.99\columnwidth}{!}{
			\begin{tabular}{l|ccc|ccc}
				\toprule
				& 			
				\multicolumn{3}{c|}{FSD} & \multicolumn{3}{c}{CenterPoint} \\
				& Mem.  & Latency(ms) & mAP & Mem.  & Latency(ms) & mAP \\
				\midrule
				all   & 5.9 & 97 & 24.0 & 10.4 & 232 & 22.0\\
				only RoI\dag  & 3.2 \lighter{45.8\%} & 81\lighter{16.5\%} & 23.2 &  9.9\lighter{4.8\%} & 227\lighter{2.2\%} & 21.5\\
				w/o ground  & 2.3 \lighter{61.0\%} & 74\lighter{25.8\%} & 21.0 & 9.7\lighter{6.7\%} & 217\lighter{6.4\%} & 19.8\\
				 \bottomrule
			\end{tabular}
		}	
	\end{center}
	\caption{
	{Performance with different detection areas.
	}
	\dag: Region of Interest is defined by the HD map in AV2 dataset.
	}
	\label{tab:map}
\end{table}
Argoverse 2 dataset provides a highly reliable HD map, which could be utilized as a prior to remove uninterested regions making the scene more sparse.
Thus we proceed with experiments removing some uninterested regions to show the advantages of FSD in more sparse scenarios.
The results are summarized in Table~\ref{tab:map}.
FSD has a significantly lower memory footprint and latency with an acceptable precision loss after removing the uninterested regions.
On the contrary, the efficiency improvement of CenterPoint is minor.
It reveals that FSD benefits more from the increase of data sparsity, which is another advantage of the fully sparse architecture.
\subsection{Comprehensive Analysis of FSD++}
\label{sec:fsdpp_analysis}
\subsubsection{Preliminary Settings for the Analysis of FSD++}
In this section, we conduct extensive experiments to reveal the inner workings of FSD++.
Here we first present the setting of our \emph{baseline} model in this section, which is slightly different from the best FSD++ model in Table.~\ref{tab:sota} and Table.~\ref{tab:sota_test}. Unless otherwise specified, the default hyper-parameters of all the FSD++ models in Sec.~\ref{sec:fsdpp_analysis} are listed in the first column of Table~\ref{tab:hyperparameter}. 
\begin{table}[h]
	\begin{center}
		\resizebox{0.99\columnwidth}{!}{
			\begin{tabular}{lccc}
				\toprule
			
				 & Baseline FSD++ & Best FSD++ & Multi-frame FSD  \\
				\midrule
				Schedule & 6 epochs & 12 epochs & 6 epochs \\
				SPS\dag & Random & Random & - \\
				\#. frames & 6 & 7 & 6\\
                Max age & 2 & 2 & - \\
                Backbone$\ast$ & SpUNet-base & SpUNet-large & SpUNet-base\\
                \#. layers in SIR2 & 3 & 3 & 3\\
                RPP size & (0.25, 0.25, 0.4) & (0.25, 0.25, 0.4) & -\\                
				 \bottomrule
			\end{tabular}
		}	
	\end{center}
		\caption{Basic hyper-parameter choice of models adopted in this section (Sec.~\ref{sec:fsdpp_analysis}) \dag: SPS stands for skeleton point sampling. $\ast$: SpUNet-large has one more stage than SpUNet-base~\cite{parta2} and the number of channels of its first is doubled. }
	\label{tab:hyperparameter}
\end{table}
\par
The model latency reported in this section is measured on a single RTX 3090 GPU with a mini-batch size of 1 in float32 precision.
To ensure accuracy, we only consider the latency of the model in all evaluations, excluding the latency of IO, which is potentially unstable in the multi-frame setting.
\par
It is worth noting that we have observed run-to-run variation in the performance of cyclist class, likely due to its low number in the dataset. As a result, we mark the performance of this class in gray in some experiments to indicate that it may not be reliable.

\subsubsection{Skeleton Point Sampling}
Table~\ref{tab:skeleton} shows the performance with different skeleton point sampling strategies.
We find that there are no significant differences between the three strategies considered.
However, using random skeleton sampling considerably reduces the latency of FSD++ without sacrificing performance.
And skeleton sampling consistently boosts the performance of the cyclist class, which suggests that appropriate sampling alleviates the overfitting for rare classes.
The performance of pedestrian is better without sampling, which reveals that more points might be helpful for small objects.
In practice, although different sampling strategies could be adopted for different classes, we use the random sampling for all classes for simplicity and generality.

\begin{table}[H]
	\begin{center}
		\resizebox{0.99\columnwidth}{!}{
			\begin{tabular}{l|cccc|c}
				\toprule
			
				&\multicolumn{4}{c}{L2 3D AP/APH} &  \multirow{2}{*}{\shortstack[1]{Latency\\(ms)}} \\
				& Mean & Vehicle & Pedestrian & Cyclist & \\
				\midrule
				Random &76.10/74.73& 72.20/71.74 & 76.93/74.11 & {79.20/78.33} & 68.7\\
				Object FPS  &75.56/74.21 & 72.06/71.59 & 76.94/74.14 & {77.68/76.88} & 72.3\\
				Voxel Sampling &75.76/74.40&  72.10/71.66 & 76.80/74.05 & {78.37/77.49} & 71.4 \\
				None &75.23/73.91& 71.93/71.50 & 77.54/74.72 & {76.33/75.51} & 73.9\\

				 \bottomrule
			\end{tabular}
		}	
	\end{center}
		\caption{Effectiveness of different skeleton point sampling.}
	\label{tab:skeleton}
\end{table}

\subsubsection{Different Number of Frames}
FSD++ samples skeleton points from multiple previous frames. Table~\ref{tab:num_frames} showcases how the number of used frames affects its performance.
There are two interesting findings.
\begin{itemize}
    \item Performance becomes better as the number of frames grows. In the meantime, the latency does not significantly increase. We owe the credit to residual point probing, which removes most of the background. It could offer even more clean residual points if more base frames (Eqn.~\ref{eqn:rpp}) are used.
    \item FSD++ outperforms FSD with the same number of frames in vehicle and cyclist class. 
    We also intuitively owe it to RPP since it removes most background clutter and eases the burden of the segmentation.
    The slightly lower performance of pedestrian suggests it might be better to retain all points for pedestrian.
    However, the performance loss is acceptable since FSD++ achieves better average performance and much lower latency.
\end{itemize}

\begin{table}[H]
	\begin{center}
		\resizebox{0.99\columnwidth}{!}{
			\begin{tabular}{l|cccc|c}
				\toprule
			
				\multirow{2}{*}{\#. frames} &\multicolumn{4}{c}{L2 3D AP/APH} &  \multirow{2}{*}{\shortstack[1]{Latency\\(ms)}} \\
				& Mean & Vehicle & Pedestrian & Cyclist &\\
				\midrule
				2 & 73.39/71.83 & 69.54/69.12 & 74.68/71.35 & \ut{75.94/75.02} & 66.1 \\
				3 &75.20/73.74& 70.95/70.52 & 76.13/73.09 & \ut{78.51/77.62} & 67.0\\
				4 &75.44/74.03& 71.67/71.21 & 76.48/73.50 & \ut{78.18/77.37} & 67.3\\
				5 &75.13/73.72& 71.50/71.04 & 76.71/73.83 & \ut{77.17/76.29} & 68.7\\
				6 &76.10/74.73& 72.19/71.74 & 76.92/74.11 & \ut{79.20/78.33} & 68.7\\
        		6 (FSD)\dag &75.65/74.28& 71.54/71.07 & 78.04/75.22 & \ut{77.37/76.54} & 116.2\\
				 \bottomrule
			\end{tabular}
		}	
	\end{center}
		\caption{Performance of FSD++ with the different number of frames. Since FSD++ uses previous foreground points, it needs at least two frames. \dag: multi-frame FSD model with simple point concatenation. Performance is unstable in the scarce cyclist class, so we mark the numbers in gray.}
	\label{tab:num_frames}
\end{table}

\subsubsection{Drifting Analysis}
It would be a major concern if FSD++ suffers from the drifting error given its reliance on history predictions. 
In particular, if the detector makes inaccurate predictions at time step $t$, it is likely that the detector becomes worse at time step $t+1$ since the predictions in $t+1$ rely on the predictions from $t$ (skeleton point sampling). 

To prevent potential drifting, we insert some keyframes at regular intervals during the inference of a sequence.
At keyframes, we use the predictions from standard FSD for skeleton point sampling, which could be viewed as a rectification of the potential drifting.
Table.~\ref{tab:drifting} shows that FSD++ achieves competitive results without any keyframes.  And the minor gap between the first row and the last row confirms that the drifting of FSD++ is negligible.
\begin{table}[H]
	\begin{center}
		\resizebox{0.99\columnwidth}{!}{
			\begin{tabular}{l|cccc}
				\toprule
			
				\multirow{2}{*}{\shortstack[1]{Gap between \\ key frames}} &\multicolumn{4}{c}{L2 3D AP/APH}  \\
				& Mean & Vehicle & Pedestrian & Cyclist \\
				\midrule
				5  &76.05/74.67& 72.23/71.77 & 76.83/74.02 & 79.08/78.21 \\
				10  &76.03/74.66& 72.20/71.75 & 76.88/74.07 & 79.00/78.15 \\
				20  &76.07/74.68& 72.20/71.74 & 76.90/74.08 & 79.11/78.23 \\
				50  &76.10/74.72& 72.20/71.74 & 76.91/74.10 & 79.20/78.32 \\
				None  &76.10/74.73& 72.20/71.74 & 76.93/74.11 & 79.20/78.33 \\
				 \bottomrule
			\end{tabular}
		}	
	\end{center}
		\caption{The performance of different keyframe gaps. ``None'' means using only initial predictions.}
	\label{tab:drifting}
\end{table}

\subsubsection{Change Blindness Ablation}
\label{sec:exp_blindness}
Due to the \emph{change blindness} we discussed in Sec.~\ref{sec:change_blindness}, newly emerged objects might be ignored by the detector.
\emph{Max age} is proposed to mitigate the issue of change blindness, and Table~\ref{tab:max_ages} shows its effect. We find keep residual points for two time steps (max age 2) is enough. 

\begin{table}[H]
	\begin{center}
		\resizebox{0.99\columnwidth}{!}{
			\begin{tabular}{l|cccc|c}
				\toprule
			
				\multirow{2}{*}{Max age} &\multicolumn{4}{c}{L2 3D AP/APH} &  \multirow{2}{*}{\shortstack[1]{Latency\\(ms)}} \\
				& Mean & Vehicle & Pedestrian & Cyclist &\\
				\midrule
				1 &75.17/73.82& 71.38/70.94 & 76.74/73.97 & \ut{77.40/76.56} & 65.3\\
				2  &76.10/74.73& 72.20/71.74 & 76.93/74.11 & \ut{79.20/78.33} & 68.7\\
				3  &\textbf{76.14}/\textbf{74.74}& \textbf{72.22}/\textbf{71.75} &  \textbf{77.06}/\textbf{74.20} & \ut{79.13/78.27} & 70.2\\

				 \bottomrule
			\end{tabular}
		}	
	\end{center}
		\caption{Different max ages of residual points. Performance is unstable in the scarce cyclist class, so we mark the numbers in gray.}
	\label{tab:max_ages}
\end{table}

For a closer look at the issue of change blindness, we split the objects in the original WOD validation set to \emph{emerging  objects} and \emph{existing objects} for further ablations.
Emerging objects mean those objects do not appear in the first frame of a sequence, while emerging later. 
The results are shown in Table~\ref{tab:blindness}.
The emerging objects recall of FSD++(1)\footnote{The numbers in the parenthesis denote the max ages. } is inferior to FSD\_6f with the same number of frames.
This suggests that change blindness is indeed an issue for FSD++. 
However, prolonging the max age makes FSD++ outperform FSD\_6f in all classes, which demonstrates the proposed max age effectively mitigates change blindness.
\begin{table}[H]
	\begin{center}
		\resizebox{0.9\columnwidth}{!}{
			\begin{tabular}{l|cccc}
				\toprule
			
				\multirow{2}{*}{Method} &\multicolumn{4}{c}{Recall of emerging objects}  \\
				& Mean & Vehicle & Pedestrian & Cyclist \\
				\midrule
				FSD &73.05& 66.01 & 72.62 & 80.53 \\
				FSD\_6f$^\ast$  &77.54& 69.74 & 78.34 & 84.53  \\
				FSD++(1) &75.58& 69.18 & 77.48 & 83.09\\
				FSD++(2)  &77.82& 70.56 & 78.10 & 84.79\\
				FSD++(3)  &78.14& 70.60 & 78.30 & 85.52\\
				 \bottomrule
			\end{tabular}
		}	
	\end{center}
		\caption{Performance for emerging objects. $\ast$: FSD with 6-frame concatenated input. In the WOD validation split, the number of emerging objects count for around 42.4\%/37.1\%/52.6\% in all objects for vehicle / pedestrian / cyclist, respectively.}
	\label{tab:blindness}
\end{table}

\subsubsection{Robustness to Seed Quality}
During inference of every point cloud sequence, FSD++ needs the predictions in the initial frame as a seed to start, as we discussed in Sec.~\ref{sec:fsdpp_inference}.
Here we figure out how the quality of seed predictions affects the performance.
Concretely, we add two typical kinds of random noise to seed predictions, including random box drop and random box insertion.
They are designed to simulate false negatives and false positives.
All the experiments share a trained FSD++ detector. The modifications above are applied during inference and are not adopted for training-time augmentation.
\par
Table~\ref{tab:seed_quality} shows FSD++ is robust to both two types of noises.
Particularly, for box drop, there is only marginal performance degradation even after dropping all the initial seed boxes.
We explain this surprising phenomenon in two aspects:
(1) FSD++ is born to be robust to the dropping of moving objects.
This is because moving objects create a considerable amount of residual points and FSD++ is capable of making predictions from these residual points.
(2) There is still a small number of residual points in the static objects due to the change of viewpoint. Moreover, the mechanism of max age also helps accumulate residual points on static objects.
\par
In the case of box insertion, FSD++ is almost unaffected because they can be easily identified as background in the segmentation stage of FSD.

\begin{table}[H]
	\begin{center}
		\resizebox{0.99\columnwidth}{!}{
			\begin{tabular}{l|cccc}
				\toprule
			
				\multirow{2}{*}{Noise type} &\multicolumn{4}{c}{L2 3D AP/APH}  \\
				& Mean & Vehicle & Pedestrian & Cyclist \\
				\midrule
				None           &76.10/74.73& 72.20/71.74 & 76.93/74.11 & 79.20/78.33 \\
				Drop (10\%)\dag & 75.95/74.58 & 72.11/71.66 & 76.81/73.99 & 78.92/78.08 \\
				Drop (50\%) &75.76/74.39& 71.86/71.41 & 76.63/73.82 & 78.78/77.95 \\
    			Drop (100\%) &74.69/73.35& 70.47/70.02 & 75.41/72.69 & 78.20/77.33 \\
				Insertion (10\%) &76.00/74.62& 72.16/71.70 & 76.82/73.99 & 79.01/78.16 \\
    			Insertion (50\%) &76.02/74.64& 72.14/71.69 & 76.86/74.04 & 79.05/78.19 \\
           	Insertion (100\%) &75.98/74.61& 72.16/71.71 & 76.84/74.02 & 78.95/78.10 \\

				 \bottomrule
			\end{tabular}
		}	
	\end{center}
		\caption{Robustness to the noisy seed predictions. \dag: the percentage in parentheses denotes the ratio of dropped/inserted instances. }
	\label{tab:seed_quality}
\end{table}

\subsubsection{Analysis of Residual Point Probing}
Quantization size and the number of base frames are two important hyper-parameters in RPP. Here we show how they affect the output residual points and performance.
\par
\noindent \textbf{Quantization size} makes RPP robust to small point disturbance. We list the results of different quantization sizes in Table~\ref{tab:rpp_size}.
It could be seen from the ``residual point ratio'' that RPP with larger quantization sizes leads to less residual points making the detector more efficient but leading to slightly lower performance.
\par
\begin{table}[H]
	\begin{center}
		\resizebox{0.99\columnwidth}{!}{
			\begin{tabular}{l|ccc|c|c}
				\toprule
			
				\multirow{2}{*}{\shortstack[1]{Quantization\\ size}} &\multicolumn{3}{c|}{L2 3D AP/APH} &  \multirow{2}{*}{\shortstack[1]{Residual \\point ratio\dag}} & \multirow{2}{*}{\shortstack[1]{Latency\\(ms)}} \\
				&  Vehicle & Pedestrian & Cyclist & &\\
				\midrule
				(0.15, 0.15, 0.4) & 72.06/71.60 & 77.19/74.43 & \ut{77.53/76.67} & 17.4\% & 72.3\\
				(0.25, 0.25, 0.4)  & 72.20/71.74 & 76.93/74.11 & \ut{79.20/78.33} & 9.6\% & 68.7\\
				(0.35, 0.35, 0.4)  & 72.04/71.58 & 76.88/74.03 & \ut{78.54/77.68} & 6.0\% & 65.2\\
				 \bottomrule
			\end{tabular}
		}	
	\end{center}
		\caption{The effectiveness of quantization size in Residual Point Probing. \dag: residual point ratio means the average ratio of the residual points to the total points in a single frame.}
	\label{tab:rpp_size}
\end{table}
\noindent \textbf{Base frame} (in Eqn.~\ref{eqn:rpp}) also has a considerable effect on RPP.
The more base frames are incorporated, the less residual points could be obtained, leading to higher efficiency. Moreover, the performance is hardly affected by the increase of base frames.
\begin{table}[H]
	\begin{center}
		\resizebox{0.99\columnwidth}{!}{
			\begin{tabular}{l|ccc|c|c}
				\toprule
			
				\multirow{2}{*}{\shortstack[1]{\#. RPP base \\ frames}} &\multicolumn{3}{c|}{L2 3D AP/APH} &  \multirow{2}{*}{\shortstack[1]{Residual \\point ratio\dag}} & \multirow{2}{*}{\shortstack[1]{Latency\\(ms)}} \\
				&  Vehicle & Pedestrian & Cyclist & &\\
				\midrule
				3 & 72.48/72.03 & 77.16/74.35 & \ut{77.60/76.78} & 14.4\% & 70.2\\
				4 & 72.24/71.79 & 77.04/74.20 & \ut{78.44/77.60} & 10.8\% & 69.0\\
				5 & 72.20/71.74 & 76.93/74.11 & \ut{79.20/78.33} & 9.6\% & 68.7\\
				6 & 72.22/71.77 & 77.41/74.59 & \ut{78.60/77.74} & 9.4\% & 68.5\\

				 \bottomrule
			\end{tabular}
		}	
	\end{center}
		\caption{The effectiveness of the number of base frames in Residual Point Probing. \dag: residual point ratio means the average ratio of the residual points to the total points in a single frame.}
	\label{tab:rbp_frames}
\end{table}

\subsection{Detailed Runtime Evaluation}
\label{sec:runtime}
\begin{figure}[h]
	\centering
	\includegraphics[width=\columnwidth]{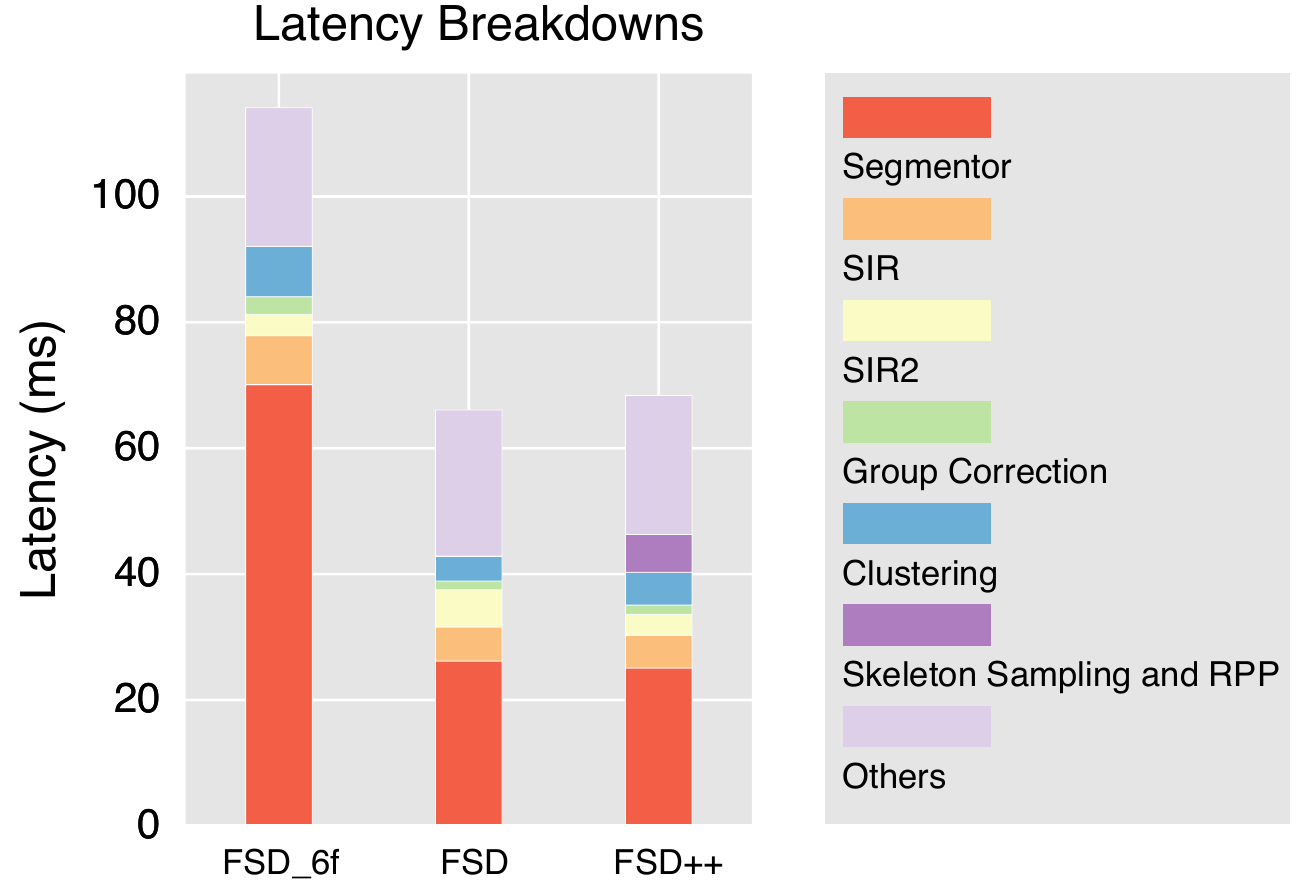}
	\caption{
	Latency breakdowns of multi-frame FSD, single-frame FSD and FSD++.}
	\label{fig:latency}
\end{figure}

Here we elaborate on the efficiency of each component of FSD and FSD++.
All evaluated models use SpUNet-large as the backbone.
Evaluations are conducted on a single RTX 3090 in FP32 precision without any test-time optimizations.
We only record the \emph{single-sample} forward latency of the detector implemented with MMDetection3Dv0.15, ignoring the IO of point clouds which is unstable in the multi-frame setting.
Fig.~\ref{fig:latency} shows the detailed results, which are average numbers evaluated on the first ten sequences of validation split.
\par
As can be seen from the figure, the latency of the segmentor is greatly reduced, which consists of the sparse voxel encode (i.e., backbone) and segmentation head.
As a results, FSD++ is as fast as the single-frame FSD, yet achieves better performance than FSD\_6f (Table~\ref{tab:num_frames}). It is worth emphasizing that the ``others'' part of latency is usually brought by some serialized operations, such as class-wise detection heads and class-wise NMS. This part of latency could be greatly reduced in deployment.


\section{Conclusion}
This paper first proposes a LiDAR-based fully sparse 3D object detection framework, namely FSD.
FSD forsakes the widely adopted dense BEV feature map in previous arts, which is the hindrance to making detectors fully sparse.
Instead, FSD consists of a general sparse voxel encoder and a highly-efficient sparse instance recognition (SIR) module.
SIR remedies the issue of the center feature missing, which is the essential difficulty of fully sparse architecture.
FSD not only actualizes efficient long-range (up to 200 meters) detection on Argoverse 2 dataset, but also achieves state-of-the-art performance on the competitive Waymo Open Dataset.
\par
To unleash the potential of FSD, we propose leveraging temporal information to remove data redundancy.
The proposed Skeleton Point Sampling and Residual Point Probing offer FSD a super sparse input point cloud, which constitutes the FSD++ framework. FSD++ achieves state-of-the-art single-model performance on both validation and test split Waymo Open Dataset, and maintains high efficiency. We hope our work lightens future research direction for LiDAR-based point cloud recognition.


%

\appendices
\section{Efficient Dynamic Pooling}
The proposed SIR consists of three basic operations: MLP, dynamic broadcasting, and dynamic pooling.
MLP is highly optimized in mainstream deep learning frameworks.
Dynamic broadcasting is essentially an indexing operation, which is highly parallel in modern GPUs.
The efficiency bottleneck of SIR lies in dynamic pooling.
Thus we provide an efficient implementation of dynamic pooling in this section.
\label{sec:appendix}
\subsection{Implementation}

The dynamic pooling implemented by PyTorch is known as scatter operation.
In this implementation, each thread manages one feature and performs simple atomic operations for feature reduction.
Intensive atomic operations in large groups are detrimental to parallelism. 

\begin{figure}[h]
	\centering
	\includegraphics[width=0.99\columnwidth]{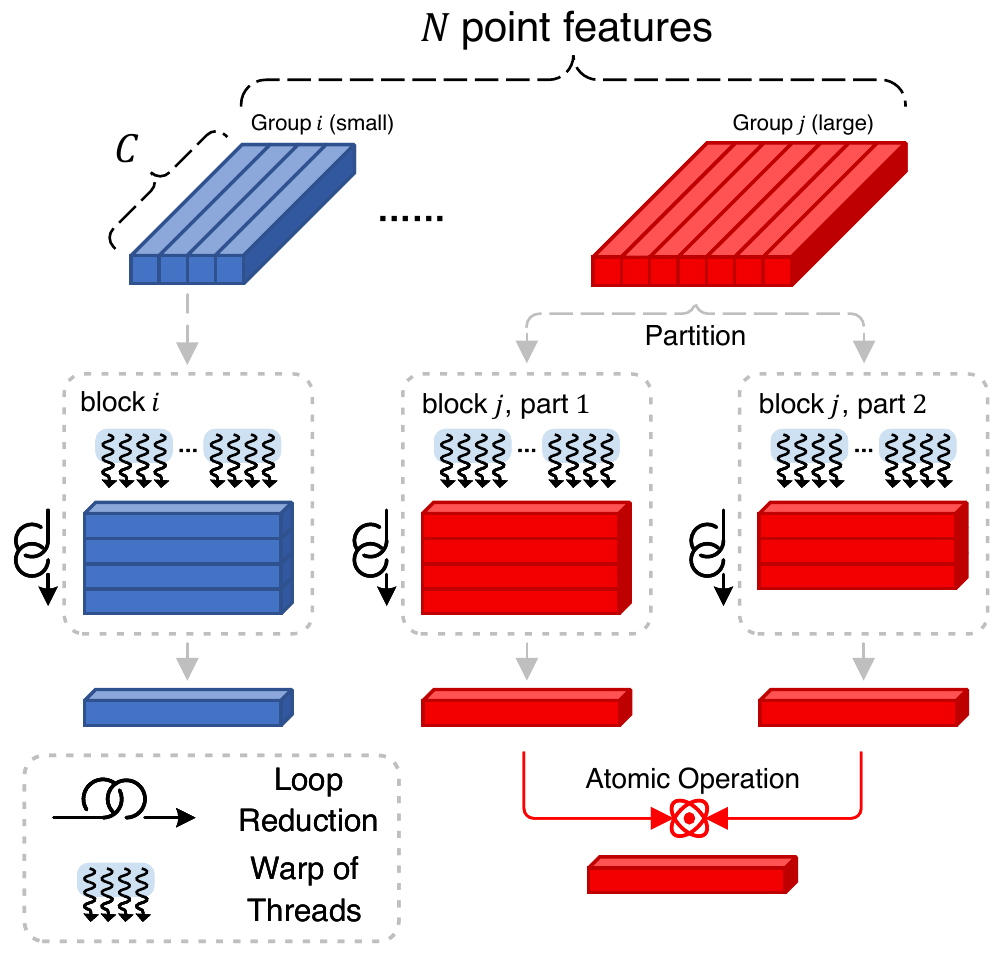}
	\caption{
	{Illustration of dynamic pooling implementation in CUDA.}
	Best viewed in color. We take two groups with different sizes as examples. Dynamic pooling reduces each group to a single feature.}
	\label{fig:scatter}
\end{figure}

We optimize the dynamic pooling operator in three aspects.
(1) Partitioning large groups into small sub-groups with fixed sizes to balance the workload.
(2) Sorting the features with the same group ID in adjacent positions to enhance the memory locality.
Note that the group IDs of each element remain unchanged throughout the SIR module, so the sort is only applied once.
(3) Threads in a warp are assigned to adjacent channels for coalesced memory access without warp divergence.
\par
In particular, each thread corresponds to a feature dimension and reduces features in a partitioned sub-group through loops.
Besides, the calls to atomic operators are reduced from once per thread to once per block, which contributes to high parallelism. 
Fig.~\ref{fig:scatter} illustrates our efficient implementation.

\subsection{Runtime Evaluation}
We take dynamic max-pooling as an example and evaluate the latency of our implementation and \textit{torch\_scatter} in different cases, including multiple feature dimensions, multiple group sizes, and whether the group size is balanced.
The total number of groups in the evaluation is 100.
The results are shown in Table~\ref{tab:balance} and Table~\ref{tab:imbalance}.
With different data sizes, our implementation achieves 2.48$\times$ to 39.58$\times$ speedup. And we have a more significant speedup with imbalanced group sizes.
\begin{table}[h]
\small
\centering

\resizebox{0.95\columnwidth}{!}{
\begin{tabular}{l|cccc}
\toprule
\multirow{2}{*}{\shortstack[1]{Feature\\ dimension}} & \multicolumn{4}{c}{Latency (ms) with Different Group Sizes} \\
 & $[10^0, 10^1)$ & $[10^1, 10^2)$ & $[10^2, 10^3)$ & $[10^3, 10^4)$\\

\midrule

64      & 0.06/0.16 & 0.06/0.16 & 0.08/1.49 & 0.24/14.05 \\
256     & 0.06/0.18 & 0.06/0.33 & 0.14/3.52 & 0.72/20.53 \\
1024    & 0.09/0.16 & 0.10/0.85 & 0.37/6.53 & 4.82/65.28 \\
\midrule
Speedup    & 2.48$\times$ & 5.56$\times$ & 20.47$\times$& 30.53$\times$ \\

\bottomrule
\end{tabular}
}
\vspace{3mm}
\caption{
Latency of dynamic max pooling on data with  \textbf{balanced} group sizes.
Each item denotes the latency of \textbf{ours/torch\_scatter} in milliseconds.
}
    \label{tab:balance}
\end{table}

\begin{table}[h]
\small
\centering

\resizebox{0.95\columnwidth}{!}{
\begin{tabular}{l|cccc}
\toprule
\multirow{2}{*}{\shortstack[1]{Feature\\ dimension}} & \multicolumn{4}{c}{Latency (ms) with Different Group Sizes} \\
 & $[10^0, 10^1)^\ast$ & $[10^1, 10^2)^\ast$ & $[10^2, 10^3)^\ast$ & $[10^3, 10^4)^\ast$\\

\midrule

64      & 0.06/0.16 & 0.06/0.32 & 0.09/2.60 & 0.36/20.87\\
256     & 0.06/0.15 & 0.08/0.78 & 0.20/5.44 & 1.19/32.75\\
1024    & 0.06/0.24 & 0.12/1.36 & 0.57/9.24 & 5.91/196.4\\
\midrule
Speedup    & 3.05$\times$ & 8.81$\times$ & 24.01$\times$ & 39.58$\times$ \\

\bottomrule
\end{tabular}
}
\vspace{3mm}
\caption{
Latency of dynamic max pooling on data with \textbf{imbalanced} group sizes.
Each item denotes the latency of \textbf{ours/torch\_scatter} in milliseconds.
$^\ast$: The sizes of one-tenth groups are enlarged by 10$\times$ to create imbalanced data.}
\label{tab:imbalance}
\end{table}




\ifCLASSOPTIONcaptionsoff
  \newpage
\fi



%


{
\bibliographystyle{IEEEtran}
\bibliography{egbib}
}

%






\end{document}